 \documentclass[final,5p,times,twocolumn,authoryear]{elsarticle}

\usepackage{amssymb}
\usepackage{lipsum}
\usepackage{booktabs}
\usepackage{float}
\usepackage{stfloats}
\usepackage{amsmath}
\usepackage[table]{xcolor}
\usepackage{colortbl}
\usepackage{caption}
\usepackage{threeparttable}

\usepackage[colorlinks=true,
            linkcolor=blue, 
            citecolor=blue,
            urlcolor=blue]{hyperref}

\journal{Elsevier}

\begin{document}

\begin{frontmatter}



\title{Trustworthy Data-Driven Wildfire Risk Prediction and Understanding in Western Canada}


\author[1]{Zhengsen Xu}
\author[2]{Lanying Wang}
\author[3]{Sibo Cheng}
\author[4]{Xue Rui}
\author[5]{Kyle Gao}
\author[1]{Yimin Zhu}
\author[1]{Mabel Heffring}
\author[1]{Zack Dewis}
\author[1]{Saeid Taleghanidoozdoozan}
\author[1]{Megan Greenwood}
\author[1,6]{Motasem Alkayid}
\author[1]{Quinn Ledingham}
\author[7]{Hongjie He}
\author[3,5,7]{Jonathan Li}
\author[1]{Lincoln Linlin Xu\corref{cor1}}
\ead{lincoln.xu@ucalgary.ca}
\cortext[cor1]{Corresponding author.}

{\raggedright
\affiliation[1]{organization={\raggedright\ignorespaces Department of Geomatics Engineering, University of Calgary},
            addressline={2500 University Drive NW}, 
            city={Calgary},
            postcode={T2N1N4}, 
            state={Alberta},
            country={Canada}}
            
\affiliation[2]{organization={Department of Environment Management, University of Waterloo},
            addressline= {200 University Avenue West},
            city={Waterloo}, 
            postcode={N2L3G1},
            state={Ontario}, 
            country={Canada}}
            
\affiliation[3]{organization={CEREA, ENPC, EDF R\&D, Institut Polytechnique de Paris},
            addressline={6--8 Avenue Blaise-Pascal, Cité Descartes},
            city={Champs-sur-Marne},
            postcode={77455},
            country={France}}

\affiliation[4]{organization={School of Emergency Management, Nanjing University of Information Science and Technology},
            addressline={219 Ningliu Road},
            city={Nanjing},
            postcode={210044},
            country={China}}

\affiliation[5]{organization={Department of Systems Design Engineering, University of Waterloo},
            addressline= {200 University Avenue West},
            city={Waterloo}, 
            postcode={N2L3G1},
            state={Ontario}, 
            country={Canada}}

\affiliation[6]{organization={Department of Geography, Faculty of Arts, The University of Jordan},
            addressline={Queen Rania Street},
            city={Amman},
            postcode={11942},
            country={Jordan}}

\affiliation[7]{organization={Hinton STAI Institute, East China Normal University},
            addressline={500 Dongchuan Road},
            city={Shanghai},
            postcode={200241},
            country={China}}
}

\begin{abstract}
In recent decades, the intensification of wildfire activity in western Canada has resulted in substantial socio-economic and environmental losses. Accurate wildfire risk prediction is hindered by the intrinsic stochasticity of ignition and spread and by nonlinear interactions among fuel conditions, meteorology, climate variability, topography, and human activities, challenging the reliability and interpretability of purely data-driven models. We propose a trustworthy data-driven wildfire risk prediction framework based on long-sequence, multi-scale temporal modeling, which integrates heterogeneous drivers while explicitly quantifying predictive uncertainty and enabling process-level interpretation. Evaluated over western Canada during the record-breaking 2023 and 2024 fire seasons, the proposed model outperforms existing time-series approaches, achieving an F1 score of 0.90 and a PR-AUC of 0.98 with low computational cost. Uncertainty-aware analysis reveals structured spatial and seasonal patterns in predictive confidence, highlighting increased uncertainty associated with ambiguous predictions and spatiotemporal decision boundaries. SHAP-based interpretation provides mechanistic understanding of wildfire controls, showing that temperature-related drivers dominate wildfire risk in both years, while moisture-related constraints play a stronger role in shaping spatial and land-cover-specific contrasts in 2024 compared to the widespread hot and dry conditions of 2023. Data and code are available at \url{https://github.com/SynUW/mmFire}.\end{abstract}




\end{frontmatter}




\section{Introduction}
\label{introduction}

Wildfire is a fundamental ecological process shaping vegetation reproduction and succession across diverse landscapes \citep{keane2010evaluating, pausas2022fire}, as well as the maintenance of habitat health \citep{keane2008ecological, pausas2019wildfires}. Consequently, it constitutes an essential component of the functioning of many ecosystems worldwide \citep{scottFireEarthIntroduction2014, wibbenmeyer2023economic}. However, wildfires also impose substantial environmental and socio-economic costs \citep{lanet2025substantial}, including reductions in carbon sequestration \citep{marshall2024can}, threats to biodiversity \citep{santos2022beyond}, emissions of hazardous gases \citep{weheba2024respiratory}, destruction of human settlements, and loss of life \citep{papathoma2022wildfire}.

Since 2000, the frequency, duration, intensity, and carbon emissions associated with wildfires in Canada have increased markedly, with particularly severe impacts observed in western provinces such as British Columbia (BC) and Alberta, where extreme fire events have become increasingly common \citep{cunningham2024increasing, mulverhill2024wildfires, christianson2024wildland, kolden2024wildfires, essd-17-5377-2025}. Between 2003 and 2022, BC experienced on average 1,350 wildfires annually, burning approximately 284,000 ha per year \citep{taylor_2022, daniels20242023}. In 2023, an unprecedented fire season occurred, driven by the combined effects of early snowmelt, prolonged drought, and persistent positive geopotential height anomalies across western Canada \citep{daniels20242023, jain2024drivers, lanet2025substantial}. A total of 2,245 wildfires burned over 15 million ha of land, forcing the evacuation of 48,000 residents and causing more than \$ 720 million in insured losses and \$ 1 billion in firefighting expenditures \citep{BC_Wildfire_Service_2023}. Given the increasing flammability of landscapes under climate change, there is an urgent need to develop advanced wildfire risk prediction approaches to enhance the socio-ecological resilience to wildfire disturbance \citep{mcwethyRethinkingResilienceWildfire2019, ager2019wildfire}.

Wildfire is inherently a stochastic phenomenon arising from nonlinear interactions among multiple driving factors, including fuel conditions, meteorological variability, topography, and human activities, operating across a wide range of spatial and temporal scales \citep{hantson2016status, kondylatos2022wildfire}. Owing to the intrinsic randomness of wildfire behavior, the absence of observed fire events does not necessarily imply the absence of fire danger, which makes quantitative modeling of wildfire risk particularly challenging \citep{prapas2021deeplearningmethodsdaily, kondylatos2022wildfire, prapas2025televit10teleconnectionawarevisiontransformers}. To reduce the complexity associated with modeling these nonlinear interactions while emphasizing the dominant influence of meteorological conditions, especially high temperature, low humidity, and strong winds, fire weather based forecasting approaches such as the Canadian Forest Fire Weather Index (FWI) system remain the most widely adopted tools for wildfire risk assessment \citep{vanwagner1987fwi, Zachary_2018, abatzoglou2019global, chenComparativeAnalysisEnsemble2025}. Although such empirical index models are easy to compute and apply, they often yield overly broad hazard conditions and fail to accurately identify potential ignition areas. Moreover, the absence of detailed fuel information frequently leads to a significant underestimation of fire severity \citep{di2025global}.

With advances in wildfire monitoring and observation, particularly the widespread use of remote sensing data, fuel conditions can now be characterized with greater accuracy, providing a foundation for higher-resolution wildfire prediction \citep{garcia2019assessment, abdollahi2023forest, xu2025deep}. At the same time, the integration of large volumes of remote sensing, meteorological, and ancillary datasets has accelerated a paradigm shift toward data-driven wildfire risk modeling \citep{kondylatos2022wildfire, jain_review_2020}. Current wildfire risk prediction studies can be broadly categorized into two classes: event-level wildfire spread prediction and large-scale wildfire ignition risk prediction \citep{xu2025bcwildfirelongtermmultifactordataset}. The former focuses on modeling the evolution and spread of an already ignited wildfire event under given environmental and meteorological conditions \citep{cheng2022data, NEURIPS2023_ebd54517, 10149031, zhao2025ts, zhao2025near, yu2025probabilistic}, whereas the latter, analogous to fire danger rating systems such as the FWI, characterizes the likelihood of wildfire ignition and potential spread at regional scales. This study falls into the second category; therefore, the term wildfire risk hereafter refers specifically to large-scale wildfire ignition risk and its associated potential for subsequent spread.

At present, data-driven wildfire risk prediction primarily relies on traditional machine learning methods, such as Random Forest and XGBoost, as well as time-series models including LSTM and ConvLSTM, which have consistently shown superior performance and generalization capability compared to conventional statistical and physically based models \citep{xu2025bcwildfirelongtermmultifactordataset, xu2025deep}. For example, \citet{di2025global} conducted a global-scale daily wildfire risk prediction task at a 9 km spatial resolution using multiple machine learning models and demonstrated that incorporating meteorological, fuel-related, topographic, and weather variables enables accurate short-term prediction of large wildfire events. Importantly, they showed that fuel-aware data-driven approaches can effectively suppress false alarms relative to FWI and better capture severe burning associated with anomalous fuel conditions. Similarly, \citet{kondylatos2022wildfire} integrated weather, fuel, topography, and human activity factors to predict next-day wildfire danger across the Mediterranean region, finding that LSTM and ConvLSTM outperform traditional machine learning methods, while all data-driven models exceed FWI in predictive accuracy. Additional studies have also confirmed the advantage of LSTM-based temporal modeling over purely spatial approaches in wildfire risk prediction \citep{deng2025daily, bhowmik_multi-modal_2023}.

Beyond conventional machine learning and standard LSTM architectures, relatively few studies have developed customized deep learning models specifically tailored for wildfire risk prediction. For instance, \citet{li2024projecting} proposed a framework that integrates LSTM-based temporal modeling with explicit representations of fuel availability, fuel combustibility, and anthropogenic fire suppression to refine wildfire risk estimates. Their results demonstrated clear advantages over traditional machine learning methods and physically based indices, including FWI, Energy Release Component (ERC), and Burning Index (BI). It is important to note that these studies primarily focus on regional-scale wildfire risk prediction, such as at the provincial level, rather than event-specific wildfire spread simulations \citep{huot2022next, zhou2025comparative}.

Taken together, these studies suggest that time-series–based approaches outperforms FWI and traditional machine learning methods. However, most existing models rely on generic temporal forecasting architectures, such as LSTM, that are not specifically designed to capture the distinctive dynamics of wildfire processes. Wildfire drivers typically exhibit pronounced multi-scale behavior, involving both long-term patterns related to fuel accumulation and climatic conditions, as well as short-term variability driven by episodic droughts and fire events \citep{jolly2015climate, touma2022climate, kondylatos2022wildfire, xu2025deep}. While wavelet-based multi-scale denoising techniques can partially disentangle these components by decomposing time-series data into distinct frequency bands while preserving temporal localization, their application in wildfire risk prediction remains limited \citep{192463, ganesan2004wavelet, harrou2024enhancing}. In addition, commonly used tree-based models and LSTM-based approaches lack explicit mechanisms for modeling interactions among heterogeneous drivers, particularly the coupling between historical fire activity and environmental conditions, as well as interactions among different environmental variables.

Furthermore, although deep learning–based wildfire risk prediction has made substantial progress, its reliability, trustworthiness, and interpretability remain major barriers to practical deployment, largely due to inherent uncertainty and limited decision transparency \citep{persello2022deep}. First, existing models are affected by data acquisition noise, epistemic uncertainty arising from limited model knowledge, and aleatoric uncertainty stemming from the intrinsic randomness of wildfire occurrence \citep{kendall2017uncertainties, lang2022global}. These sources of uncertainty can propagate through the prediction process, leading to erroneous risk estimates and, consequently, unreliable guidance for decision-making \citep{zhang2025calibration}.


Research on uncertainty quantification in wildfire risk prediction remains relatively limited. Among existing efforts, \citet{kondylatos2025uncertaintyawaredeeplearningwildfire} were among the first to propose a unified Bayesian deep learning framework that jointly estimates epistemic and aleatoric uncertainty. In this framework, epistemic uncertainty is approximated using Bayesian neural networks trained via Bayes by Backprop (BBB), Monte Carlo Dropout, and Deep Ensembles, while aleatoric uncertainty is modeled by imposing a heteroscedastic Gaussian distribution over the softmax logits following \citet{collier2020simpleprobabilisticmethoddeep}. The two uncertainty components are jointly estimated during inference through a double Monte Carlo sampling procedure to derive total predictive uncertainty. Although this study systematically assessed the effectiveness of multiple uncertainty estimation strategies and demonstrated the feasibility of disentangling epistemic and aleatoric uncertainty, it did not investigate the spatiotemporal structure of predictive uncertainty or its dependence on environmental context.

\begin{figure*}[b]
    \centering
    \includegraphics[width=1\linewidth]{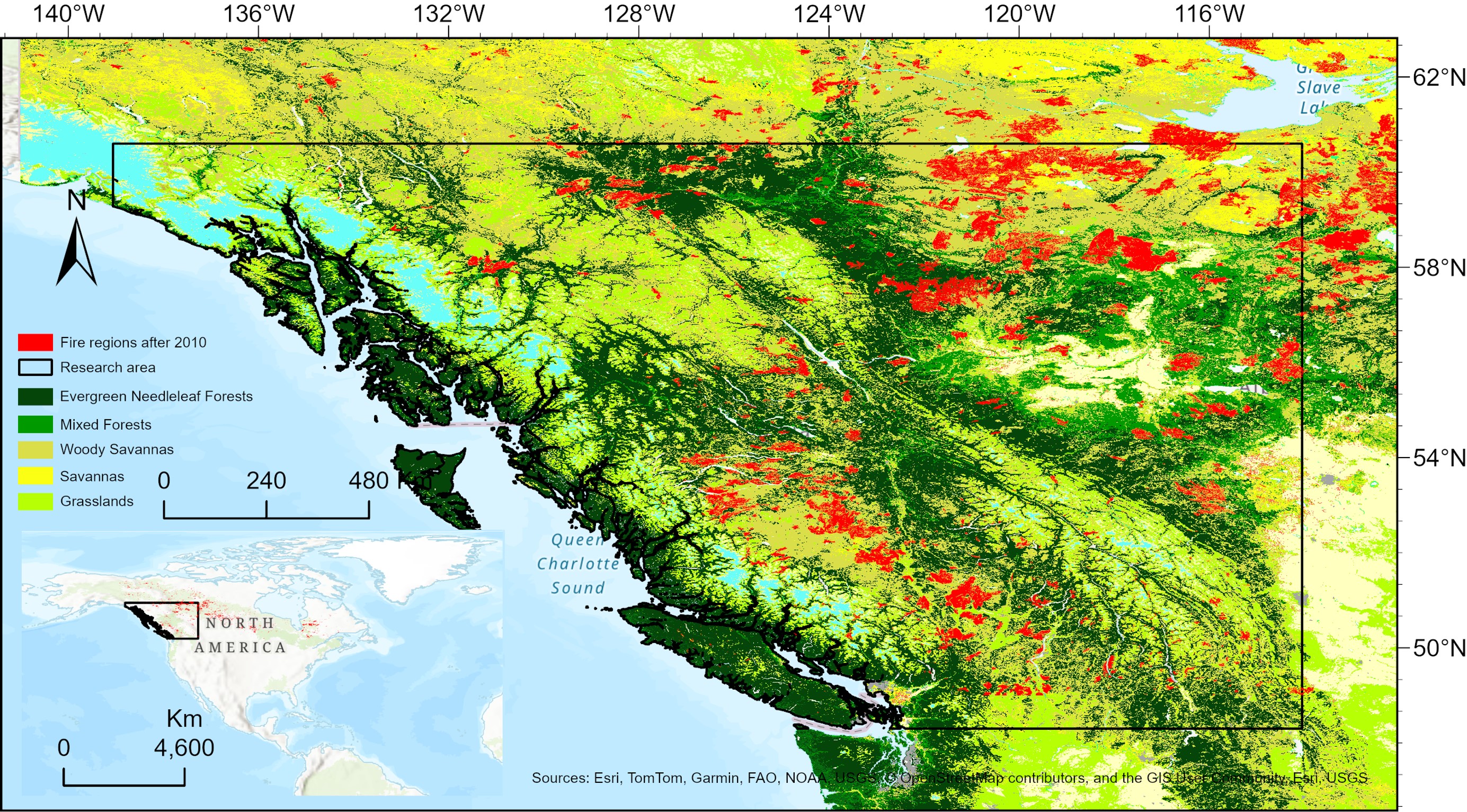}
    \caption{Map of research area overlaid with burned areas since 2010 from the Canadian Wildland Fire Information System and dominate land-cover types.}
    \label{fig:research_area}
\end{figure*}

On the other hand, because wildfire occurrence is governed by the complex interactions among historical fires, weather, fuel, topography, and human activity, it is not only essential to predict fire occurrence accurately but also to identify and understand the dominant driving factors that control wildfire dynamics for effective fire management \citep{wang2021identifying, li2024projecting}. To enhance the transparency of deep learning–based decision-making, explainable artificial intelligence (xAI) has become an increasingly important tool in Earth observation and wildfire research \citep{gevaert2022explainable, ishikawa2023example, li2024projecting, jiang2024interpretable}. xAI methods provide post hoc interpretability by quantifying both the magnitude and direction (positive or negative) of each variable’s influence on model predictions \citep{arrieta2020explainable, li2024projecting}.

Despite their promise, xAI applications in deep learning–based wildfire modeling remain limited, with most studies focusing on wildfire risk prediction in the United States \citep{li2024projecting} and the Mediterranean region \citep{kondylatos2022wildfire}. For instance, \citet{li2024projecting} applied SHapley Additive exPlanations (SHAP) \citep{NIPS2017_8a20a862} to evaluate the marginal contribution of each feature to wildfire predictions, including both the magnitude and direction of influence. Building upon SHAP analysis, \citet{kondylatos2022wildfire} further employed Partial Dependence Plots (PDP) \citep{molnar2020interpretable} and Integrated Gradients (IG) \citep{sundararajan2017axiomatic} to investigate how variations in both the value and temporal evolution of individual drivers affect wildfire risk. In addition, several studies using traditional machine learning models have incorporated SHAP-based analysis to quantify feature-level marginal contributions \citep{wang2021identifying, pelletier_wildfire_2023}.

Although existing studies have successfully integrated interpretability methods with deep learning frameworks, most of them implicitly assume spatial homogeneity within the study area and do not account for differences in wildfire risk attribution across distinct fuel or land-cover types \citep{pozo2022assessing, yang2023characterization, 10.1071/WF24025}. In addition, prior wildfire xAI research has predominantly focused on the United States and Mediterranean regions, with comparatively limited attention given to western Canada. Western Canada is characterized by substantial carbon storage, diverse ecosystems, and frequent wildfire activity, making it a particularly important yet underexplored region for deep learning wildfire risk modeling and analysis \citep{xu2025bcwildfirelongtermmultifactordataset}.

In summary, current deep learning–based wildfire risk prediction faces three major challenges. First, there is a lack of deep learning architectures explicitly tailored for wildfire risk prediction that adequately capture multi-scale dynamics and complex interactions among multiple driving factors, resulting in suboptimal predictive performance. Second, existing wildfire risk prediction models often lack uncertainty quantification, undermining their trustworthiness and reliability. Third, fine-grained analyses of wildfire-driving factors in western Canada remain scarce despite the region’s ecological and climatic significance.

To address the above limitations, this study makes the following three main contributions:

\begin{enumerate}

\item We introduce a wavelet-based multi-scale temporal modeling framework for next-day wildfire risk prediction that captures long-term dependencies and cross-feature interactions among wildfire driving factors.

\item We introduce an entropy-based uncertainty estimation strategy to enhance the trustworthiness of wildfire risk predictions, and further perform spatiotemporal decomposition of predictive uncertainty to characterize variations in uncertainty across environmental contexts and fire-season stages.

\item We disentangle SHAP-based feature attribution results across different years and land-cover types to analyze the marginal contributions of wildfire drivers at multiple spatial and temporal scales in western Canada, thereby revealing the heterogeneous mechanisms that control wildfire risk across ecosystems and climatic conditions.

\end{enumerate}

\section{Data Construction and Sampling}
\label{Data}

\begin{table*}[b]
\centering
\caption{Summary of input variables, their categories, frequency, sources, and references.}
\begin{tabular}{lllll}
\hline
\hline
\textbf{Variable} & \textbf{Input category} & \textbf{Frequency} & \textbf{Source} & \textbf{Reference} \\
\midrule
2m Temperature & Weather & Daily & ERA5-Land & \citet{munoz2021era5} \\
10m Wind Speed & Weather & Daily & ERA5-Land & \citet{munoz2021era5} \\
2m Dewpoint Temperature & Weather & Daily & ERA5-Land & \citet{munoz2021era5} \\
Surface Latent Heat Flux & Weather & Daily & ERA5-Land & \citet{munoz2021era5} \\
Snow Cover & Weather & Daily & ERA5-Land & \citet{munoz2021era5} \\
Surface Pressure & Weather & Daily & ERA5-Land & \citet{munoz2021era5} \\
Total Precipitation & Weather & Daily & ERA5-Land & \citet{munoz2021era5} \\
Volumetric Soil Water L1-L4 & Weather & Daily & ERA5-Land & \citet{munoz2021era5} \\
MOD/MYD11A1 & Weather & Daily & MODIS & \citet{wan2021mod11a1_v061} \\
MOD/MYD09CMG & Weather & Daily & MODIS & \citet{vermote2015mod09cmg} \\
MOD/MYD09GA & Fuel & Daily & MODIS & \citet{vermote2015mod09ga} \\
NDVI and EVI & Fuel & Daily & MOD/MYD09GA \\
MCD15A3H & Fuel & 4 Days & MODIS & \citet{myneni2015mcd15a3h} \\
MCD12Q1 & Fuel & Yearly & MODIS & \citet{friedl2019mcd12q1} \\
MOD/MYD14A1 & Ignition & Daily & MODIS & \citet{giglio2015mod14a1} \\
ASTER GDEM v3 & Topography & Constant & ASTER & \citet{ASTERGDEMv3_2019} \\
Water-body polygons & Topography & Constant & OSM & \citet{OpenStreetMap2025} \\
Settlement Distribution & Human Activity & Constant & OSM & \citet{OpenStreetMap2025} \\
Power Grid Distribution & Human Activity & Constant & OSM & \citet{OpenStreetMap2025} \\
\hline
\hline
\end{tabular}
\label{tab:input_variables}
\end{table*}

Our dataset covers the entire province of British Columbia, West part of Alberta, and portions of Alaska and the continental United States (\autoref{fig:research_area}), spanning the period from 2000 to 2024 and a total area of approximately 240~million hectares. This region was selected because it is characterized by rich carbon storage, frequent wildfire occurrence, and diverse yet fragile ecosystems, making it both scientifically important and highly challenging for wildfire risk prediction.

To effectively train and evaluate deep learning models, we constructed an integrated multi-drivers dataset with a spatial and temporal resolution of 1~km~$\times$~1~km~$\times$~1~day. The dataset combines historical wildfire point observations with multiple categories of wildfire-related driving factors, including weather, fuel, topography, and human activity, as shown in \autoref{tab:input_variables}. The composition of these driving factors and their corresponding preprocessing workflows are described in the following sections.

\subsection{Weather}

Meteorological conditions are the dominant determinants of wildfire behavior and can affect ignition, spread, and persistence \citep{abatzoglou2013relationships, preisler2016near, bhowmik_multi-modal_2023, deng2025daily}. High temperature, low humidity, limited precipitation, and strong wind generally increase fire risk \citep{zacharakis2023integrated, xu2025deep}. Accordingly, we used twelve variables from the ERA5-Land reanalysis dataset, including 2~m temperature, 10~m~U and V wind components, snow cover, 24-hour total precipitation, surface latent heat flux, 2~m dew point temperature, surface pressure, and volumetric soil water content for four soil layers \citep{munoz2021era5}. Following \citet{kondylatos2022wildfire}, we used daily aggregated values to represent the meteorological conditions most conducive to fire activity. Since the native spatial resolution of ERA5-Land is approximately 11~km, we resampled it to 1~km using the nearest-neighbor method.

In contrast to studies that relied solely on reanalysis data \citep{wang2021identifying, kondylatos2022wildfire, li2024projecting, deng2025daily, kondylatos2025uncertaintyawaredeeplearningwildfire}, we additionally incorporated MODIS infrared and thermal products to describe finer-scale surface temperature dynamics. Specifically, MOD/MYD11A1 \citep{wan2021mod11a1_v061} provides 1~km day and night land surface temperature and emissivity from Bands~31 and~32, while MOD/MYD09CMG provides 5.6~km daily brightness temperature from Bands~20,~21,~31,~32. We applied quality control layers to remove low-quality observations, merged MOD and MYD data, and filled the remaining gaps with historical data to generate seamless daily composites without information leakage. Both datasets were then resampled to 1~km spatial resolution.

\begin{figure*}
    \centering
    \includegraphics[width=1\linewidth]{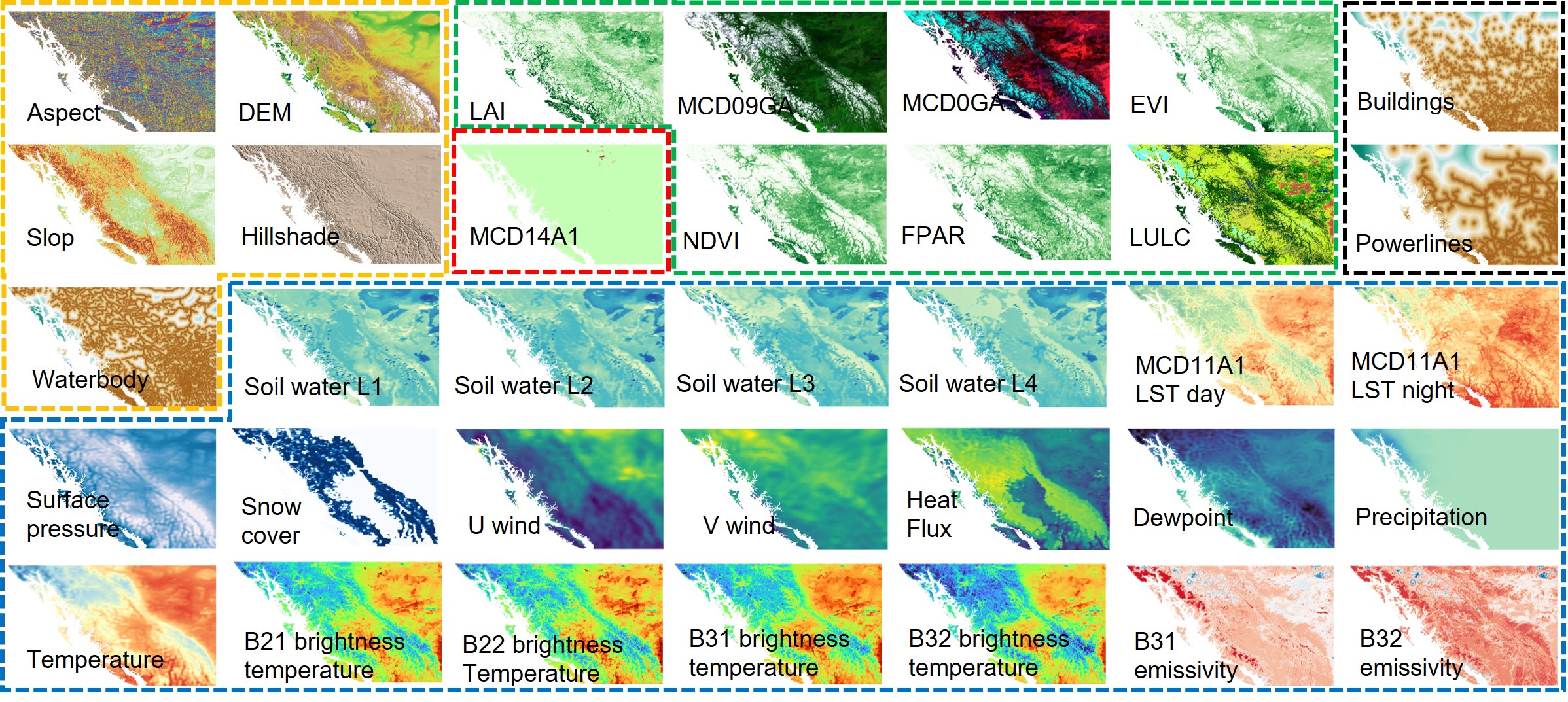}
    \caption{Partial overview of wildfire-driving factors, including fuel conditions (green), fire detection (red), topographic attributes (yellow), human activity indicators (black), and meteorological variables (blue).}
    \label{fig:dataset_overview_real}
\end{figure*}

\begin{figure*}[b]
    \centering
    \includegraphics[width=1\linewidth]{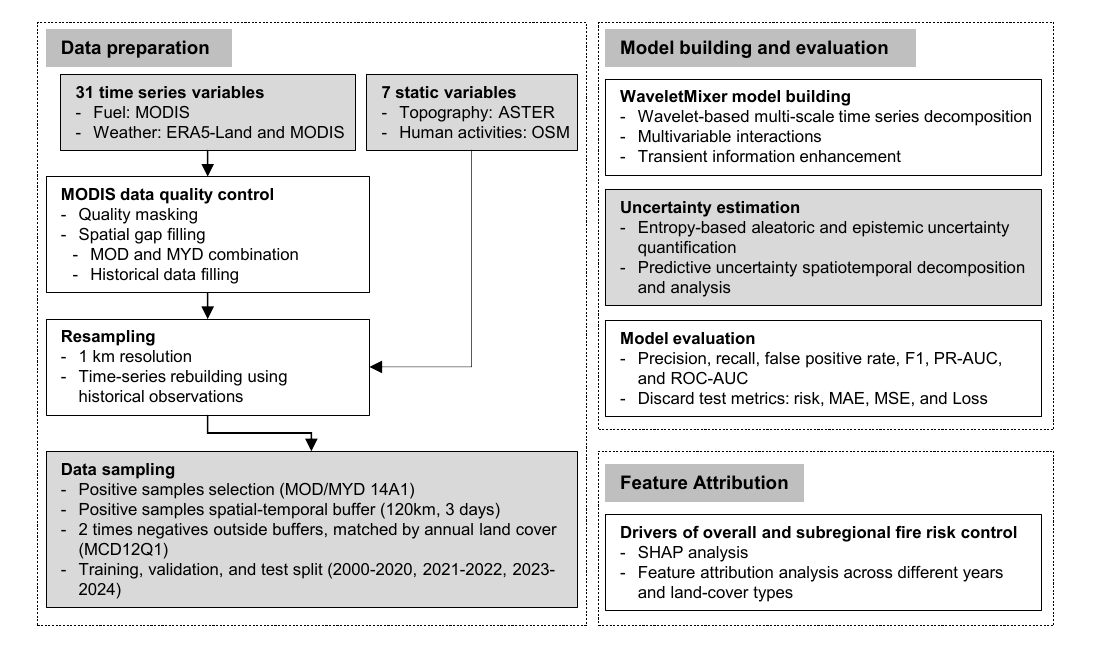}
    \caption{Conceptual diagram of the data preparation, model building, uncertainty estimation, probability calibration, model performance evaluation and multi-scale drivers of fire risk control.}
    \label{fig:overall_diagram}
\end{figure*}

\subsection{Fuel}

Wildfire behavior depends strongly on the fuel that sustains combustion, and fuel characteristics are among the most manageable wildfire drivers \citep{thompson2013quantifying, duff2017revisiting}. However, direct quantification of fuel load, fuel type, and fuel continuity is difficult and prone to cumulative uncertainty. Therefore, remote sensing–based vegetation indices are often used as proxies for fuel conditions \citep{xu2025deep}. We used MODIS products, including daily MOD/MYD09GA, 4-day MCD15A3H, and annual MCD12Q1, to represent vegetation and fuel status.

The MOD/MYD09GA product includes visible and near-infrared bands (Bands~1,~2,~3, and~7). We applied the same preprocessing procedure as for MOD/MYD11A1 and MOD/MYD09CMG, which includes low-quality filtering, MOD–MYD merging, seamless daily construction, and 1~km resampling. We further calculated NDVI and EVI to indicate vegetation greenness and vigor. The MCD15A3H product provides 4-day composite LAI and FPAR data, which were gap-filled using historical records and replicated to daily frequency. The annual MCD12Q1 land cover data were also replicated to daily intervals for temporal consistency. These indices together provide a reliable proxy for vegetation health, fuel load, and fuel continuity.

\subsection{Historical Fire Data}

Historical ignition records serve as both labels for model training and as dynamic variables that reflect the influence of past fire events on subsequent risk. To ensure long-term and consistent coverage, we used the MOD/MYD14A1 active fire product. Low-quality and low-confidence detections (confidence~\textless~8) were removed using QA layers, and pixels with confidence~$\geq$~8 were assigned a value of~1 to represent active fire. MOD and MYD data were then merged so that if either indicated a fire, the merged dataset retained the value~1; otherwise, the value was set to~0.

\subsection{Topography}

Topographic conditions influence fuel properties and local microclimate \citep{xu2025deep}. Elevation regulates temperature and humidity, thereby affecting vegetation type and decomposition rate. Aspect determines solar exposure and evapotranspiration, which influence fuel moisture and structure, while slope affects soil water retention and the rate of fire spread \citep{carmo2011land, calvino2017interacting}. To further describe moisture and terrain effects, we included a water-body proximity layer, representing the distance to the nearest water body. 

We obtained the DEM from ASTER GDEM~v3 \citep{ASTERGDEMv3_2019}, aggregated to 1~km by mean resampling, and derived slope, aspect, and hillshade. Water-body polygons were obtained from OpenStreetMap (OSM) and rasterized to 1~km resolution, and distances to the nearest water bodies were computed for all locations \citep{OpenStreetMap2025}.

\subsection{Human Activity}

Human activities such as agricultural burning, arson, or other open-fire events can trigger wildfires \citep{deng2025daily}. In contrast, fire suppression and management may reduce fire occurrence. Infrastructure such as power lines can also act as ignition sources \citep{9409078}. To account for anthropogenic influences, we included three spatial variables: distance to roads, distance to powerlines, and distance to settlements. We rasterized OSM-based settlement, powerline, and road layers to 1~km resolution and computed Euclidean distances to the nearest features \citep{OpenStreetMap2025}. As topography and human activity layers were static, they were replicated across all days to ensure temporal consistency.


\subsection{Data Sampling}

The overall data preparation can be see in \autoref{fig:overall_diagram} and partial drivers can be see in \autoref{fig:dataset_overview_real}. We formulated wildfire prediction as a binary time-series forecasting task, predicting whether fire occurs within a given future window. Fire pixels in MOD/MYD14A1 dataset were labeled as positive samples and non-fire pixels as negative samples. Since wildfires are rare events, the dataset was highly imbalanced. Following prior studies \citep{kondylatos2023mesogeos, kondylatos2025uncertaintyawaredeeplearningwildfire}, we applied stratified sampling with a positive-to-negative ratio of~1:2.

Considering that areas near recent fires remain high-risk, we excluded data within 60~km and~$\pm$3~days around each fire event when selecting negative samples. To avoid trivial learning from land-cover differences, we enforced equal land-cover type distribution between positive and negative samples within each year, maintaining a 1:2 ratio in every land-cover category. This design balances class representation and ensures meaningful contrast between fire and non-fire conditions.

\section{Methodology}

To develop a trustworthy data-driven framework for deep learning–based wildfire risk prediction, we design an integrated methodology that encompasses model construction, uncertainty estimation, and factor-importance analysis, as illustrated in \autoref{fig:overall_diagram}. First, we introduce the proposed WaveletMixer architecture for long-sequence wildfire risk forecasting, which serves as a foundation for representing the complex temporal dependencies inherent in wildfire drivers. Building upon this model, we address limitations in predictive trustworthiness by incorporating an entropy-based uncertainty estimation strategy to quantify predictive uncertainty and support more reliable risk interpretation. Finally, SHAP-based analyses are employed to assess the contributions of individual environmental and anthropogenic drivers, providing process-level insights into the mechanisms governing wildfire risk dynamics.

\begin{figure*}[b]
    \centering
    \includegraphics[width=1\linewidth]{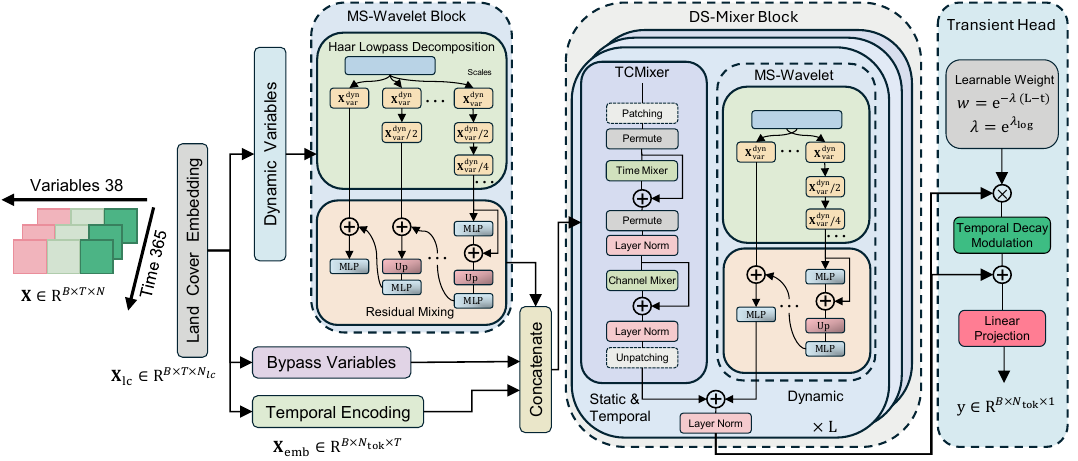}
    \caption{The overall architecture of the proposed Multi-Scale Wavelet--Fusion Forecasting Model. \textcircled{+}, and \textcircled{$\times$} represent concatenation and multiply, respectively.}
    \label{fig:model_architecture}
\end{figure*}

\subsection{Multi-Scale Wavelet Fusion Mixer with Dynamic–Static Separation}

This subsection introduces the Multi-Scale Wavelet Fusion Mixer with Dynamic–Static Separation (WaveletMixer), a sequence-to-sequence architecture designed to accommodate the heterogeneous temporal properties of wildfire-related drivers. As illustrated in \autoref{fig:model_architecture}, the model operates on daily multivariate sequences and processes the inputs through three main stages.

Given an input tensor $\mathbf{X}\in\mathbb{R}^{B\times T\times N}$, the model first applies a dedicated embedding layer to transform the categorical land-cover variable into a continuous latent representation. The resulting feature tensor $\mathbf{X}{\mathrm{emb}}\in\mathbb{R}^{B\times T\times N_{tok}}$ which also contains calendar-time encodings $\mathbf{x}{\mathrm{mark}}\in\mathbb{R}^{B\times T\times N_{\mathrm{mark}}}$, is then partitioned into dynamic and static subsets based on their temporal characteristics.


Dynamic variables, such as fuel and meteorological factors, exhibit pronounced multi-scale characteristics, encompassing both long-term patterns and short-term fluctuations. Wavelet-based multi-scale denoising methods have been shown to effectively disentangle signal components across multiple temporal scales \citep{192463, ganesan2004wavelet, 6522407, harrou2024enhancing}. Accordingly, dynamic variables are first passed through a multi-scale Haar wavelet module (MS-Wavelet Block). This module performs hierarchical temporal decomposition and residual mixing to separate the driving data into frequency-specific components, selectively filtering noise and spurious variability while preserving salient features and patterns across multiple scales. In contrast, static variables and deterministic calendar-time features bypass this operation and are directly routed to the Dynamic–Static Multi-Scale Mixer (DS-Mixer Block).


Within the Dynamic--Static Multi-Scale Mixer (DS-Mixer), a dual-branch structure is applied iteratively for $L$ layers, reflecting distinct inductive biases for dynamic and static inputs. Dynamic variables are processed through the MS-Wavelet Block at each layer, where wavelet-based denoising leverages a cascade of scaling and wavelet functions to progressively approximate the intrinsic dynamics of the driving variables across multiple temporal scales. In contrast, static variables are modeled through a combination of patch-wise hidden-space mixing and inter-channel mixing, without imposing explicit temporal structure. The outputs of the two branches are fused additively and normalized at each layer, enabling joint representation learning across dynamic and static feature spaces.

Finally, the fused representation is passed through a transient exponential-weighting head, which modulates recent temporal information, followed by a linear projection layer to produce the forecasted sequence. Each component is detailed below.

\subsubsection{Land Cover Embedding and Temporal Encoding}

For each spatial location, we construct a daily multivariate sequence of length $T$, where the $t$-th day is represented by a feature vector $\mathbf{x}_t \in \mathbb{R}^{N}$, $t=1,\dots,T$. The input drivers include continuous variables such as meteorological, fuel, topographic, and distance-based indicators, together with a categorical land-cover channel. The land-cover variable is treated as a discrete label $c_t \in \{1,\dots,17\}$. Since these categorical values have no inherent ordinal structure, we separate the land-cover channel from the continuous drivers and encode it using a learnable embedding.

Let $k_t$ denote the sanitized land-cover index at time $t$. A learnable embedding matrix
\begin{equation}
\mathbf{E}_{\mathrm{lc}} \in \mathbb{R}^{K_{\mathrm{lc}} \times N_{\mathrm{lc}}},
\end{equation}
maps $k_t$ to the embedding vector
\begin{equation}
\mathbf{e}^{\mathrm{lc}}_t = \mathbf{E}_{\mathrm{lc}}[k_t] \in \mathbb{R}^{N_{\mathrm{lc}}}.
\end{equation}

In addition to the environmental drivers, we construct a calendar-time feature tensor $\mathbf{x}_{\mathrm{mark}}$, consisting of seven deterministic encodings describing annual progression (e.g., normalized month, day, and weekday using sinusoidal functions). Because these covariates are deterministic functions of calendar time rather than realizations of a stochastic environmental process, they are incorporated directly asauxiliary channels with other static variables and are not subjected to multi-scale temporal decomposition.

Finally, the augmented feature at time $t$ is obtained by concatenating the continuous 
drivers, the land-cover embedding, and the calendar-time encodings:

\begin{equation}
\tilde{\mathbf{x}}_t
=
\bigl[
\mathbf{x}^{\mathrm{cont}}_t,\;
\mathbf{e}^{\mathrm{lc}}_t,\;
\mathbf{x}_{\mathrm{mark},\,t}
\bigr]
\in \mathbb{R}^{N_{emb}},
\quad
N_{tok} = N_{\mathrm{cont}} + N_{\mathrm{lc}} + N_{\mathrm{mark}}.
\end{equation}

\subsubsection{Dynamic and Static Channel Partitioning}

For a batch of size $B$, the preprocessed driver sequence is arranged as $\mathbf{X}_{\mathrm{emb}} \in \mathbb{R}^{B \times T \times N_{tok}}$. We partition the feature channels into three groups: a set of dynamic variables that exhibit meaningful temporal evolution, a set of bypass variables corresponding to static or wildfire detection (sparse) channels that are not passed through the multi-scale dynamic module, and the temporal covariates from $\mathbf{x}_{\mathrm{mark}}$. Let $N_{\mathrm{dyn}}$, $N_{\mathrm{byp}}$, and $N_{\mathrm{mark}}$ denote the numbers of dynamic, bypass, and time-covariate channels, respectively, and $N_{\mathrm{tok}} = N_{\mathrm{dyn}} + N_{\mathrm{byp}} + N_{\mathrm{mark}}$ the total number of channels used in the fusion stage. Using fixed index sets, we extract $\mathbf{X}_{\mathrm{dyn}} \in \mathbb{R}^{B \times T \times N_{\mathrm{dyn}}}$ and $\mathbf{X}_{\mathrm{byp}} \in \mathbb{R}^{B \times T \times N_{\mathrm{byp}}}$ from $\mathbf{X}_{\mathrm{emb}}$, and later concatenate them with the temporal covariates $\mathbf{x}_{\mathrm{mark}}$ along the feature dimension when entering the fusion blocks.  


\subsubsection{Multi-Scale Wavelet Block for Dynamic Drivers}

Wildfire-related dynamic drivers, such as weather and fuel indicators, exhibit entangled temporal behavior across heterogeneous time scales: ignition-critical day-to-day anomalies are superimposed on slowly evolving seasonal and climatic backgrounds. This multi-scale entanglement makes it difficult for a single-resolution representation to simultaneously capture short-term precursors and long-term conditioning. To address this, and to capture temporal patterns at multiple resolutions while preserving the original sequence length, we introduce a Multi-scale wavelet block that applies recursive Haar low-pass filtering to decompose dynamic channels. These coarse-to-fine representations are fused via a bottom-up residual pathway, yielding a length-preserving sequence that synthesizes both global trends and local transients.

The input to this block is $\mathbf{X}_{\mathrm{dyn}} \in \mathbb{R}^{B \times T \times N_{\mathrm{dyn}}}$, which is first permuted to $\mathbf{U}^{(0)} \in \mathbb{R}^{B \times N_{\mathrm{dyn}} \times T}$, so that low-pass filtering operates along the temporal dimension. At each scale $s=0,\dots,S-1$, a Haar low-pass operator is applied pairwise along time. The low-frequency component at the next scale is then computed as
\begin{equation}
\ell^{(s+1)}_{b,c,t}
=
\frac{1}{2}
\left(
U^{(s)}_{b,c,2t}
+
U^{(s)}_{b,c,2t+1}
\right),
\end{equation}
where $b$ indexes the batch, $c$ the channel, and $t$ the time index at the coarser scale.  
This yields $\mathbf{U}^{(s+1)} \in \mathbb{R}^{B \times N_{\mathrm{dyn}} \times L_{s+1}}$ with $L_{s+1} = \lfloor L_s/2 \rfloor$. Repeating this operation $S$ times produces a collection of representations $\{\mathbf{U}^{(s)}\}_{s=0}^{S}$ at progressively coarser temporal resolutions.

At the coarsest scale $S$, a fully connected network $f^{(S)}$ operates on $\mathbf{U}^{(S)}$ to refine large-scale trend features:
\begin{equation}
\tilde{\mathbf{U}}^{(S)}
=
\mathbf{U}^{(S)} + f^{(S)}\bigl(\mathbf{U}^{(S)}\bigr),
\end{equation}
where $f^{(S)}$ is a two-layer multilayer perceptron with hidden dimension $d_{\mathrm{hid}}$, ReLU activation, and dropout.  
The refined coarsest representation is then propagated back to finer scales in a bottom-up manner.  
For each scale $s = S-1,\dots,0$, the coarser representation $\tilde{\mathbf{U}}^{(s+1)}$ is upsampled along time to match the length of $\mathbf{U}^{(s)}$ using a linear interpolation operator denoted by $\mathcal{I}_{\uparrow}(\cdot)$.  
A scale-specific MLP $f^{(s)}$ produces a residual correction:
\begin{equation}
\tilde{\mathbf{U}}^{(s)}
=
\mathbf{U}^{(s)}
+
f^{(s)}\Bigl(
    \mathcal{I}_{\uparrow}\bigl(\tilde{\mathbf{U}}^{(s+1)}\bigr)
\Bigr),
\end{equation}
where $f^{(s)}$ shares the same structure as $f^{(S)}$ but with independent parameters.  
The finest-scale output $\tilde{\mathbf{U}}^{(0)}$ is finally permuted back to shape $B \times T \times N_{\mathrm{dyn}}$ and used as the multi-scale refined dynamic representation $\tilde{\mathbf{X}}_{\mathrm{dyn}}$.

It is worth noting that the MS-Wavelet Block is invoked at two distinct stages of the model. First, the block is applied directly to the raw dynamic drivers at the input stage, acting as a global multi-scale encoder that disentangles fast-changing ignition precursors from slowly varying environmental baselines before any higher-level temporal interaction is performed. This produces a denoised and scale-aware representation that is shared across all subsequent layers.

Second, within each DS-Mixer block (see \autoref{fig:model_architecture}), a separate instance of the MS-Wavelet Block operates on the evolving dynamic features in parallel with the PCMixer path. In this setting, the module functions as a layer-wise multi-scale refinement mechanism: it continuously injects explicit coarse-to-fine corrections into the hidden states, ensuring that multi-scale temporal structure is preserved throughout the deeper stages of the encoder. 

\subsubsection{Time–Channel Mixer}

To jointly model local temporal structure and cross-variable interactions at the native resolution, we employ a Time–Channel mixing module (TCMixer in \autoref{fig:model_architecture}). The input to this module is the concatenation of the multi-scale refined dynamic drivers, bypassed static variables, and deterministic temporal covariates:
\begin{equation}
\mathbf{X}_{\mathrm{comb}}
=
\bigl[
    \tilde{\mathbf{X}}_{\mathrm{dyn}},\;
    \mathbf{X}_{\mathrm{byp}},\;
    \mathbf{x}_{\mathrm{mark}}
\bigr]
\in \mathbb{R}^{B \times T \times N_{\mathrm{tot}}}.
\end{equation}

Directly modeling long temporal sequences at full resolution requires the network to learn both rapid short-term fluctuations and slower background variations within a single receptive field, which is inefficient and prone to overfitting. Following the intuition behind patchification in multi-scale time-series models \citep{nie2023a}, we partition the sequence into non-overlapping segments of length $P$. Each patch serves as a local temporal window in which short-range fluctuations can be modeled independently. Before patchification, boundary replication is applied to pad the sequence length to a multiple of $P$, resulting in a reshaped tensor $\mathbf{Z} \in \mathbb{R}^{B \times N_{\mathrm{patch}} \times P \times N_{\mathrm{tot}}}$, where $N_{\mathrm{patch}}$ is the number of temporal patches.

In Time Mixer, for each patch and each variable, the $P$-length temporal vector is processed using a two-layer temporal MLP:
\begin{equation}
\mathrm{MLP}_{\mathrm{time}}(\mathbf{z}) 
= 
\sigma(\mathbf{z} \times \mathbf{W}^{(1)}_{\mathrm{time}} + \mathbf{b}^{(1)}_{\mathrm{time}}) \times 
    \mathbf{W}^{(2)}_{\mathrm{time}} + \mathbf{b}^{(2)}_{\mathrm{time}},
\end{equation}
where $\mathbf{z} \in \mathbb{R}^{P}$, and $\mathbf{W}^{(1)}_{\mathrm{time}}, \mathbf{W}^{(2)}_{\mathrm{time}}$ are learnable matrices. This operation mixes local temporal patterns (e.g., humidity drops, temperature surges) within each patch. A residual connection and layer normalization are then applied along the variable dimension.

Once temporal mixing is performed, each position inside a patch corresponds to a feature vector in $\mathbb{R}^{N_{\mathrm{tot}}}$ encoding all drivers at that timestep.  To capture cross-variable relationships, we apply another two-layer MLP along the feature dimension:
\begin{equation}
\mathrm{MLP}_{\mathrm{chan}}(\mathbf{v})
=
\sigma(\mathbf{v} \times \mathbf{W}^{(1)}_{\mathrm{chan}} + \mathbf{b}^{(1)}_{\mathrm{chan}}) \times
    \mathbf{W}^{(2)}_{\mathrm{chan}} + \mathbf{b}^{(2)}_{\mathrm{chan}},
\end{equation}
where $\mathbf{v} \in \mathbb{R}^{N_{\mathrm{tot}}}$. This operation enables the model to integrate multivariate information (e.g., wind--fuel coupling, humidity--temperature interactions) critical for describing wildfire-relevant environmental states. A second residual path and layer normalization stabilize optimization.

The processed patches are finally reshaped back to the original temporal resolution, and any padded elements are removed. The resulting representation preserves both fine-scale temporal cues and inter-variable dependencies while maintaining computational efficiency.

\subsubsection{Parallel Fusion Strategy}

A single-path time-series forecasting architecture must jointly handle rapidly varying dynamic drivers, slowly drifting background variables, and deterministic temporal covariates within the same computational stream. Such heterogeneous temporal behaviors impose conflicting requirements on a unified module, making it difficult for one pathway to simultaneously account for fine-scale fluctuations and broader structural patterns. To better align the model design with these disparate characteristics, we adopt a parallel fusion strategy that separates the processing of dynamic and full-feature representations.

In this fusion block, the dynamic subset of variables is routed through the multi-scale wavelet pathway, while the complete feature set is processed in parallel by the aforementioned TCMixer. Let $\mathbf{X}_{\mathrm{time}}$ denote the output of the TCMixer branch and $\mathbf{X}_{\mathrm{ms}}$ the reconstructed tensor in which only the dynamic channels have been transformed by the MS-Wavelet Block.  
The two pathways are then combined through additive coupling followed by layer normalization:
\begin{equation}
\mathbf{H}
=
\mathrm{LN}\!\left(
    \mathbf{X}_{\mathrm{time}}
    +
    \mathbf{X}_{\mathrm{ms}}
\right),
\end{equation}
where $\mathrm{LN}(\cdot)$ denotes layer normalization applied along the channel dimension.  
The fused output $\mathbf{H}$ serves as the input to the next fusion block, and this process is repeated across $L$ stacked layers.

\subsubsection{Transient Temporal Head and Forecasting Output}

Following the $L$ stacked parallel fusion blocks, the model produces an encoded representation $\mathbf{H}_{\mathrm{enc}} \in \mathbb{R}^{B \times T \times N_{\mathrm{tot}}}$, which contains both multi-scale dynamic responses and full-sequence time--channel interactions. Before projecting this representation to the prediction horizon, a transient temporal head is applied to introduce an explicit temporal weighting mechanism aligned with the sequential structure of the input.

Let $t=0,\dots,T-1$ denote the temporal index and let 
$\lambda = \exp(\lambda_{\log}) > 0$
be a learnable scalar that regulates an exponential weighting profile.  
The weight assigned to each time step is defined as
\begin{equation}
w_t
=
\exp\bigl[-\lambda\,(T-1-t)\bigr],
\end{equation}
which depends solely on the deterministic position of each step within the input window.  
Collecting these weights into a tensor 
$\mathbf{W} \in \mathbb{R}^{B \times T \times N_{\mathrm{tot}}}$
via broadcasting, the transient-adjusted representation is written as
\begin{equation}
\mathbf{H}_{\mathrm{tr}}
=
\mathbf{H}_{\mathrm{enc}}
+
\mathbf{W}\odot\mathbf{H}_{\mathrm{enc}},
\end{equation}
where $\odot$ denotes element-wise multiplication.  
This formulation preserves the encoded multivariate structure while allowing the model to assign a learnable temporal profile across the input window.

The final forecasting step is achieved by a linear projection operating along the temporal axis.  
For each channel $c$ with transient representation 
$\mathbf{h}^{(c)} \in \mathbb{R}^{T}$,
the prediction for a horizon of $T_{\mathrm{pred}}$ steps is computed as
\begin{equation}
\hat{\mathbf{z}}^{(c)}
=
\mathbf{W}_{\mathrm{proj}}\,
\mathbf{h}^{(c)}
+
\mathbf{b}_{\mathrm{proj}}
\in \mathbb{R}^{T_{\mathrm{pred}}},
\end{equation}
where $\mathbf{W}_{\mathrm{proj}} \in \mathbb{R}^{T_{\mathrm{pred}} \times T}$ and 
$\mathbf{b}_{\mathrm{proj}} \in \mathbb{R}^{T_{\mathrm{pred}}}$  
are shared across all channels.  
The predicted dynamic and bypass variables are subsequently rearranged into their original ordering, while deterministic temporal covariates are excluded from the forecast.  
The resulting output provides a multivariate sequence of $T_{\mathrm{pred}}$ steps for all physical driver channels at each spatial location.


\subsection{Uncertainty Estimation}

The abstraction of real-world perception inevitably introduces mismatches among models, data, and the underlying reality. Consequently, representing complex natural and social processes using limited data and models built upon incomplete knowledge inherently leads to uncertainty \citep{hall2003handling, zhang2025calibration}. In general, uncertainty is considered to arise from two primary sources: aleatoric uncertainty (also referred to as data uncertainty) and epistemic uncertainty (also referred to as model uncertainty) \citep{der2009aleatory, kendall2017uncertainties}. Aleatoric uncertainty reflects the intrinsic randomness and noise present in observational data, as well as the unpredictability of human activities and environmental dynamics. This type of uncertainty is unavoidable and cannot be reduced by collecting more training data. In contrast, epistemic uncertainty stems from limited understanding and incomplete knowledge of the underlying processes being modeled; it can be mitigated by acquiring additional data or by developing more expressive models \citep{ghosh2007nonparametric, lang2022global}.

Currently, the main approaches for uncertainty estimation include Bayesian methods, entropy-based methods, and variance-based methods \citep{ghosh2007nonparametric, namdari2019review}. Among them, entropy-based approaches are particularly well suited for classification problems where the posterior distribution follows a Bernoulli scheme, while also being computationally more efficient than fully Bayesian formulations \citep{zhang2025calibration}. Therefore, in this study, we adopt an entropy-based framework to decompose and quantify both epistemic and aleatoric uncertainties. Specifically, we decompose the model’s total predictive entropy into the expected conditional entropy (representing aleatoric uncertainty) and the mutual information (representing epistemic uncertainty), providing an measure of the two major sources of uncertainty.

Consider a probabilistic classifier with parameters $\omega$ and a learned posterior distribution $p(\omega \mid \mathcal{D})$ given training data $\mathcal{D}$. For a new input $\mathbf{x}$, the Bayesian predictive distribution over the label $y$ is obtained by marginalizing over model parameters:
\begin{equation}
    p(y \mid \mathbf{x}, \mathcal{D})
    = \int p(y \mid \mathbf{x}, \omega)\, p(\omega \mid \mathcal{D})\, \mathrm{d}\omega.
    \label{eq:predictive_distribution}
\end{equation}

The total predictive uncertainty for input $\mathbf{x}$ can be quantified by the predictive entropy:
\begin{equation}
    H_{\text{pred}}(\mathbf{x})
    = H\bigl[ p(y \mid \mathbf{x}, \mathcal{D}) \bigr]
    = - \sum_{y} p(y \mid \mathbf{x}, \mathcal{D}) \log p(y \mid \mathbf{x}, \mathcal{D}),
    \label{eq:predictive_entropy}
\end{equation}
where the sum runs over all discrete classes. This entropy reflects both uncertainty due to inherent data noise and uncertainty due to lack of knowledge about the model parameters.

To disentangle these contributions, we treat the model parameters $\omega$ as a random variable with posterior $p(\omega \mid \mathcal{D})$ and use the conditional entropy $H[y \mid \mathbf{x}, \omega]$ to quantify the uncertainty that remains even if $\omega$ were known exactly. The aleatoric uncertainty at $\mathbf{x}$ is then defined as the posterior expectation of this conditional entropy:
\begin{equation}
\begin{aligned}
H_{\text{data}}(\mathbf{x})
&= \mathbb{E}_{p(\omega \mid \mathcal{D})}
   \bigl[ H\bigl[ p(y \mid \mathbf{x}, \omega) \bigr] \bigr] \\[3pt]
&= \int H\bigl[ p(y \mid \mathbf{x}, \omega) \bigr]\,
   p(\omega \mid \mathcal{D})\, \mathrm{d}\omega,
\label{eq:aleatoric_definition}
\end{aligned}
\end{equation}
with
\begin{equation}
    H\bigl[ p(y \mid \mathbf{x}, \omega) \bigr]
    = - \sum_{y} p(y \mid \mathbf{x}, \omega) \log p(y \mid \mathbf{x}, \omega).
    \label{eq:conditional_entropy}
\end{equation}

The remaining part of the predictive uncertainty is attributed to uncertainty about the parameters $\omega$. This epistemic uncertainty can be expressed as the mutual information between $y$ and $\omega$ given $(\mathbf{x}, \mathcal{D})$:
\begin{equation}
    I\left[ y, \omega \mid \mathbf{x}, \mathcal{D} \right]
    = H\bigl[ p(y \mid \mathbf{x}, \mathcal{D}) \bigr]
      - \mathbb{E}_{p(\omega \mid \mathcal{D})}
        \bigl[ H\bigl[ p(y \mid \mathbf{x}, \omega) \bigr] \bigr].
    \label{eq:mutual_information_definition}
\end{equation}
Substituting \autoref{eq:predictive_entropy} and \autoref{eq:aleatoric_definition} into \autoref{eq:mutual_information_definition}, we obtain the following decomposition:
\begin{equation}
    H_{\text{pred}}(\mathbf{x})
    = H_{\text{data}}(\mathbf{x})
      + I\left[ y, \omega \mid \mathbf{x}, \mathcal{D} \right].
    \label{eq:entropy_decomposition}
\end{equation}
\autoref{eq:entropy_decomposition} shows that the total predictive entropy can be additively decomposed into a data-dependent term $H_{\text{data}}(\mathbf{x})$ (aleatoric uncertainty), which cannot be reduced by collecting more data, and a parameter-dependent term $I[y, \omega \mid \mathbf{x}, \mathcal{D}]$ (epistemic uncertainty), which can in principle be reduced by increasing the amount or diversity of training data.

In practice, the posterior distribution \(p(\omega \mid \mathcal{D})\) is intractable and thus commonly approximated via Bayesian  sampling \citep{kondylatos2025uncertaintyawaredeeplearningwildfire}. Two typical strategies are MC Dropout and Deep Ensembles, which provide stochastic realizations of model parameters that can be interpreted as approximate posterior samples \citep{pmlr-v48-gal16,Lakshminarayanan_de, kondylatos2025uncertaintyawaredeeplearningwildfire}.  
Let us denote by \(\{\omega_m\}_{m=1}^{M}\) a collection of \(M\) parameter samples drawn from \(p(\omega \mid \mathcal{D})\) or its approximation, and by
\begin{equation}
    p_m(y \mid \mathbf{x})
    \equiv p(y \mid \mathbf{x}, \omega_m)
\end{equation}
the corresponding predictive distribution produced by the \(m\)-th sampled model.

The marginal predictive distribution can then be approximated by averaging over these sampled models:
\begin{equation}
    \hat{p}(y \mid \mathbf{x}, \mathcal{D})
    = \frac{1}{M} \sum_{m=1}^{M} p_m(y \mid \mathbf{x}),
    \label{eq:mc_predictive}
\end{equation}
which serves as a Monte Carlo estimate of \autoref{eq:predictive_distribution}.  
The total predictive uncertainty is quantified by the entropy of this marginal predictive distribution:
\begin{equation}
    \widehat{H}_{\text{pred}}(\mathbf{x})
    = - \sum_{y} \hat{p}(y \mid \mathbf{x}, \mathcal{D})
          \log \hat{p}(y \mid \mathbf{x}, \mathcal{D}).
    \label{eq:mc_predictive_entropy}
\end{equation}

Similarly, the aleatoric uncertainty is estimated as the average conditional entropy over all model samples:
\begin{equation}
    \widehat{H}_{\text{data}}(\mathbf{x})
    = \frac{1}{M} \sum_{m=1}^{M}
      \left(
        - \sum_{y} p_m(y \mid \mathbf{x})
          \log p_m(y \mid \mathbf{x})
      \right),
    \label{eq:mc_aleatoric}
\end{equation}
which captures the expected intrinsic randomness of the data given a fixed model configuration.

The remaining part of the predictive entropy, obtained as the difference between the total and aleatoric components, represents the epistemic uncertainty and is equivalent to the mutual information between predictions and model parameters:
\begin{equation}
    \widehat{I}\left[ y, \omega \mid \mathbf{x}, \mathcal{D} \right]
    = \widehat{H}_{\text{pred}}(\mathbf{x})
      - \widehat{H}_{\text{data}}(\mathbf{x}).
    \label{eq:mc_epistemic}
\end{equation}

By construction, these empirical estimators satisfy the approximate decomposition:
\begin{equation}
    \widehat{H}_{\text{pred}}(\mathbf{x})
    \approx \widehat{H}_{\text{data}}(\mathbf{x})
      + \widehat{I}\left[ y, \omega \mid \mathbf{x}, \mathcal{D} \right],
\end{equation}
which serves as the Monte Carlo counterpart of the exact information-theoretic identity in \autoref{eq:entropy_decomposition}. This formulation provides a practical means to disentangle and quantify both aleatoric and epistemic components of uncertainty within deep neural networks.

While Bayesian sampling can be approximated via deep ensembles, training many models is computationally expensive \citep{lang2022global}, and MC Dropout often yields unstable estimates \citep{9150658, kondylatos2025uncertaintyawaredeeplearningwildfire}. To obtain diverse posterior samples without retraining multiple networks, we design a Checkpoint Ensemble approach inspired by Snapshot Ensembles \citep{huang2017snapshot}. During training, we store multiple checkpoints that naturally arise from the non-convex optimization trajectory of stochastic gradient descent. Each checkpoint $\omega_m$ thus provides a distinct model instance:
\begin{equation}
    \omega_m \sim \text{Checkpoints}\bigl(\omega(t_m)\bigr),
    \quad m = 1,\dots,M.
\end{equation}
To further enrich posterior diversity, we store checkpoints corresponding to different performance optima under metrics such as F1-score, recall, PR-AUC, MAE, and MSE. These complementary checkpoints form an ensemble that better reflects posterior variability. This Checkpoint Ensemble provides a practical trade-off between MC Dropout and Deep Ensembles: it requires only a single training process while still capturing multiple modes of the parameter space.

\subsection{Interpretability Analysis}

\begin{figure*}[b]
    \centering
    \includegraphics[width=1\linewidth]{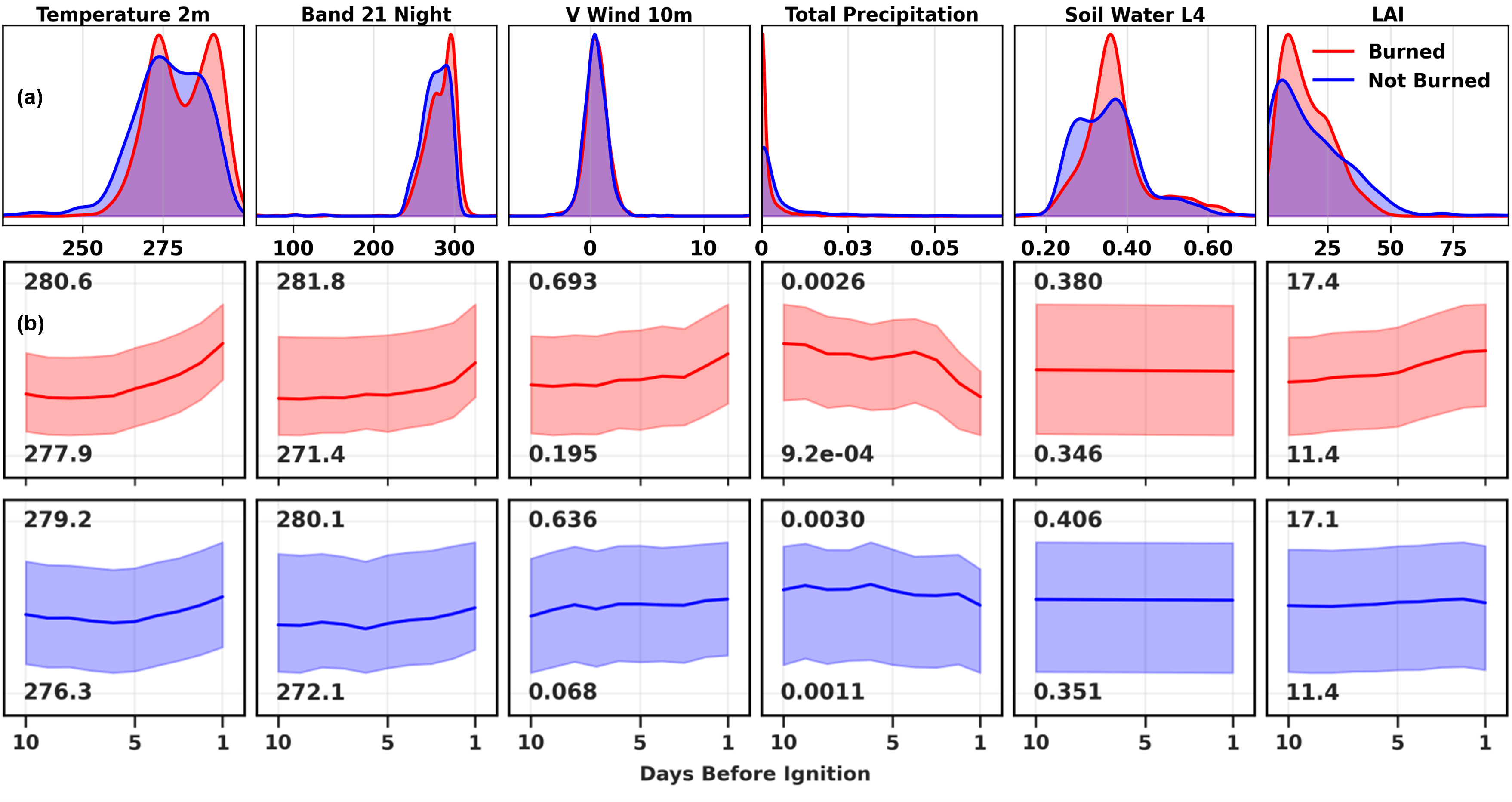}
    \caption{Distributions and temporal patterns of key environmental variables in burned and unburned locations prior to ignition. (a) Kernel density estimates for 6 selected drivers during the 10 days preceding fire events. (b) Temporal evolution of the same variables over the corresponding period.}
    \label{fig:distribution}
\end{figure*}

Interpreting wildfire risk prediction models is essential for understanding how physical processes regulate wildfire occurrence. Most existing studies rely on global SHAP-based interpretation to quantify the marginal contribution of individual driving factors~\citep{wang2021identifying, kondylatos2022wildfire, li2024projecting,deng2025daily}. While effective for identifying overall feature importance, such global analyses primarily reflect spatially averaged effects and are unable to capture the pronounced heterogeneity of wildfire-controlling mechanisms across fuel types and land-cover types. In reality, wildfire regimes vary substantially in terms of frequency, extent, intensity, and severity across fuel types and ecosystems~\citep{gallardo2016impacts, erni2020developing, gannon2021global}, implying that the relative importance and directional influence of driving factors are inherently context dependent.

To address this limitation, we build upon the proposed WaveletMixer model and develop a spatially decomposed SHAP-based feature attribution framework that enables land-cover-specific interpretation of wildfire drivers. This framework identifies the dominant controls within individual land-cover regions and explicitly reveals the regional heterogeneity of the underlying wildfire mechanisms.

Formally, for interpretability analysis, the model is trained once on the full dataset and subsequently evaluated over different land cover subsets, denoted as zones \(z \in \mathcal{Z}\). Within each zone, the model output \(p(y \mid \mathbf{x}, \omega)\) represents the predicted probability of wildfire occurrence given input feature vector \(\mathbf{x} = (x_1, x_2, \dots, x_d)\) and model parameters \(\omega\). The goal of interpretability analysis is to quantify how variations in each input feature influence this predictive probability across spatial and temporal dimensions.

For each land-cover type \(z\), the Shapley value of feature \(x_i\) is defined as:
\begin{equation}
\phi_{i,z}
= \mathbb{E}_{\mathbf{x}_{\setminus i} \sim P(\mathbf{x}_{\setminus i} \mid z)}
\big[
p(y{=}1 \mid \mathbf{x}, \omega)
-
p(y{=}1 \mid \mathbf{x}_{\setminus i}, \omega)
\big],
\label{eq:simple_shapley_zone}
\end{equation}
where \(\mathbf{x}_{S}\) denotes the subset of features indexed by \(S\), and the subscript \([\cdot]_z\) indicates that the evaluation is restricted to samples belonging to zone \(z\). This formulation quantifies the marginal contribution of feature \(x_i\) to the model’s predicted wildfire probability within each land cover subset.

\subsection{Experimental Setup}

\subsubsection{Dataset Preparation}


To prevent information leakage and preserve the temporal dependencies inherent in wildfire processes, we adopt a strictly chronological data split: observations from 2000--2020 are used for model training, data from 2021--2022 form the validation set, and data from 2023--2024 are held out for testing. This temporal partition inevitably introduces a prior shift between training and evaluation periods; however, such shifts naturally occur in real wildfire systems and therefore provide a realistic assessment of model robustness.

Beyond the challenges introduced by this prior shift, we further examine the distributions of wildfire-relevant drivers for ignition and non-ignition samples to characterize overlap and separability in the input feature space. Specifically, we analyze the empirical distributions of several key drivers using kernel density estimates (\autoref{fig:distribution} (a)), and investigate the multi-day evolution of drivers preceding ignition and non-ignition cases (\autoref{fig:distribution} (b)). \autoref{fig:distribution} (a) reveals substantial overlap between the positive and negative distributions for most variables, underscoring the intrinsic difficulty of wildfire prediction and the stochastic nature of ignition. Nevertheless, moisture- and fuel-related variables exhibit comparatively clearer separation, suggesting greater sensitivity to ignition processes. \autoref{fig:distribution} (b) further indicates that increasing temperature, reduced precipitation, atmospheric drying, and progressive fuel accumulation frequently precede wildfire occurrence.

\subsubsection{Implementation Details}

To evaluate the predictive accuracy of the proposed long-sequence wildfire risk model, we compare it against a suite of state-of-the-art deep learning baselines, including linear architectures (TSMixer \citep{chen2023tsmixer}, CrossLinear \citep{cross_linear}), Transformer-based models (Autoformer \citep{NEURIPS2021_bcc0d400}, Pyraformer \citep{liu2022pyraformer}, iTransforemr \citep{liu2024itransformer}), Mamba-based S\_Mamba \citep{wang2025mamba}, and the classical LSTM \citep{hochreiter1997long}. All models are trained, validated, and tested using identical datasets, normalization strategies, and hyperparameter settings.

All experiments are conducted using the PyTorch framework on an NVIDIA RTX A6000 GPU. Prior to training, each driver is normalized independently using min--max scaling. Models are optimized using Adam with a fixed learning rate of $1\times 10^{-5}$ over 50 epochs, with early stopping applied using a patience of 10 epochs. The loss function for all methods is the standard cross-entropy loss. 

To improve training efficiency, all blocks in the proposed WaveletMixer model are configured with two layers and a hidden dimension of 1024. The number of wavelet decomposition scales is set to 3, allowing the model to obtain progressively coarser representations through successive Haar low-pass operations.

\subsubsection{Evaluation Metrics}

Let $\mathrm{TP}$, $\mathrm{FP}$, $\mathrm{TN}$, and $\mathrm{FN}$ denote true positives, false positives, true negatives, and false negatives, respectively.  
Precision, recall, F1-score, and the false positive rate (FPR) are defined as:
\begin{equation}
    \mathrm{Precision}
    = \frac{\mathrm{TP}}{\mathrm{TP}+\mathrm{FP}}, \qquad
\end{equation}

\begin{equation}
        \mathrm{Recall}
    = \frac{\mathrm{TP}}{\mathrm{TP}+\mathrm{FN}},
\end{equation}

\begin{equation}
    \mathrm{F1} 
    = \frac{2\times \mathrm{Precision}\times \mathrm{Recall}}
           {\mathrm{Precision}+\mathrm{Recall}}, \qquad
\end{equation}

\begin{equation}
\mathrm{FPR} = \frac{\mathrm{FP}}{\mathrm{FP} + \mathrm{TN}}.
\end{equation}



Precision, recall, and the F1-score are used to evaluate the accuracy of wildfire event prediction. Notably, given the asymmetric consequences associated with wildfire risk, recall is of particular importance, as failing to identify high-risk events can lead to substantial losses. The false positive rate (FPR) is additionally reported to quantify the frequency of false alarms, which is critical for operational risk assessment. Mean squared error (MSE) and mean absolute error (MAE) are further used to assess the accuracy of probabilistic predictions. To provide a threshold-independent evaluation of classification performance, we report the area under the precision–recall curve (PR-AUC). Model efficiency is evaluated using the number of trainable parameters and floating-point operations (FLOPs), enabling a comparative assessment of computational complexity across methods.

For uncertainty-based selective prediction experiments, we adopt the notion of risk from selective classification~\citep{geifman2017selective}:
\begin{equation}
R(f,g)
=
\frac{\mathbb{E}\!\left[ \ell\!\left(f(\mathbf{x}), y\right)\, g(\mathbf{x}) \right]}
{\mathbb{E}\!\left[ g(\mathbf{x}) \right]},
\end{equation}
where \(f\) denotes the prediction model, \(g(\mathbf{x}) \in \{0,1\}\) is a selection function indicating whether a prediction is retained, and \(\ell(\cdot)\) represents the prediction loss. In practice, risk corresponds to the empirical error rate computed over the subset of samples retained after uncertainty-based rejection, while coverage is defined as the fraction of retained samples. By examining the trade-off between risk and coverage, we evaluate the effectiveness of uncertainty estimates in selectively discarding unreliable predictions while preserving stable predictive performance on the remaining samples.






\section{Results and Discussion}

\subsection{Performance on Wildfire Risk Prediction Models}

\subsubsection{Quantitative Evaluation}


\begin{table*}[b]
    \centering
    \caption{Comparison of the proposed WaveletMixer with other time-series forecasting methods on the 2023 and 2024 test sets. The best and second-best results are highlighted in purple and green. MSE and MAE are multiplied by 100.}
    \begin{tabular}{ccccccccccc}

\hline												
\hline
Model	&	Precision$\uparrow$	&	Recall$\uparrow$	&	F1$\uparrow$	&	PR\_AUC$\uparrow$	&	ROC\_AUC$\uparrow$	&	FPR$\downarrow$	&	MSE$\downarrow$	&	MAE$\downarrow$	& Param(M)$\downarrow$ & FLOPs(G)$\downarrow$\\
\hline																	
Autoformer	&	\cellcolor{green!20} \textbf{94.59}	&	63.79	&	76.19	&	91.79	&	90.01	&	\cellcolor{green!20}\textbf{3.65}	&	13.80	&	23.45	&	14.95	&	399.12 \\

Pyraformer	&	91.49	&	77.05	&	83.65	&	94.81	&	94.40	&	7.17	&	11.17	&	17.00&	7.04	&	393.22	\\

iTransformer	&	92.38	&	66.41	&	77.28	&	91.10	&	88.96	&	\cellcolor{blue!20}\textbf{5.47}	&	15.99	&	20.64 &	41.94	&	129.35	\\

S\_Mamba	&	91.88	&	78.71	&	\cellcolor{blue!20}\textbf{84.79}	&	95.00	&	94.11	&	6.95	&	10.59	&	\cellcolor{blue!20}\textbf{15.89}	&	16.97	&	23.63	\\

CrossLinear	&	88.73	&	70.96	&	78.86	&	92.63	&	91.43	&	9.01	&	13.31	&	20.14	&	30.33	&	239.74	\\

LSTM	&	91.86	&	70.96	&	80.06	&	93.51	&	92.64	&	6.29	&	13.01	&	19.32	&	1.51	&	\cellcolor{green!20}\textbf{6.71}	\\

TSMixer	&	\cellcolor{blue!20}\textbf{92.59}	&	\cellcolor{blue!20}\textbf{78.02}	&	84.68	&	\cellcolor{blue!20}\textbf{95.54}	&	\cellcolor{blue!20}\textbf{94.95}	&	6.25	&	\cellcolor{blue!20}\textbf{9.91}	&	16.10	&	\cellcolor{blue!20}\textbf{0.83}	&	\cellcolor{blue!20}\textbf{7.28}	\\

\hline
WaveletMixer	&	92.29	&	\cellcolor{green!20}\textbf{88.21}	&	\cellcolor{green!20}\textbf{90.20}	&	\cellcolor{green!20}\textbf{97.36}	&	\cellcolor{green!20}\textbf{97.15}	&	7.37	&	\cellcolor{green!20}\textbf{6.97}	&	\cellcolor{green!20}\textbf{11.81}	&	\cellcolor{green!20}\textbf{0.49}	&	16.9	\\

\hline
\hline

\end{tabular}						
    \label{tab:comparison}
\end{table*}

\autoref{tab:comparison} summarizes the performance and computational efficiency of the proposed WaveletMixer model in comparison with several state-of-the-art temporal baselines on the 2023 and 2024 test sets. The model achieves precision, recall, and F1 values close to 0.9, and its PR\_AUC and ROC\_AUC values both exceed 0.97. At the same time, its FPR, mean squared error (MSE), and mean absolute error (MAE) remain relatively low at 7.37\%, 0.0697, and 0.1181. These results indicate that WaveletMixer provides strong prediction performance at a fixed threshold and maintains reliable ranking ability and good consistency between predicted and true probabilities across the full output distribution.

Compared with all baseline models, WaveletMixer exhibits clear advantages in both predictive accuracy and computational efficiency. For example, its recall exceeds that of the second-best model by 10.19\% (88.21\% versus 78.02\%), while precision decreases by only 0.3\% (92.58\% versus 92.29\%). The F1 score improves by approximately 5.4\% relative to the S\_Mamba model. Threshold-independent metrics yield similar conclusions, since PR\_AUC and ROC\_AUC exceed those of TSMixer by 1.82\% and 2.2\%. The model also achieves the lowest MSE and RMSE among all competitors, reducing these values by 0.0294 and 0.0408 relative to the next-best method. Despite these performance gains, WaveletMixer remains lightweight, containing only 0.49M parameters and requiring 16.9G FLOPs.

\begin{figure}[t]
    \centering
    \includegraphics[
    width=0.95\columnwidth,
    ]
    {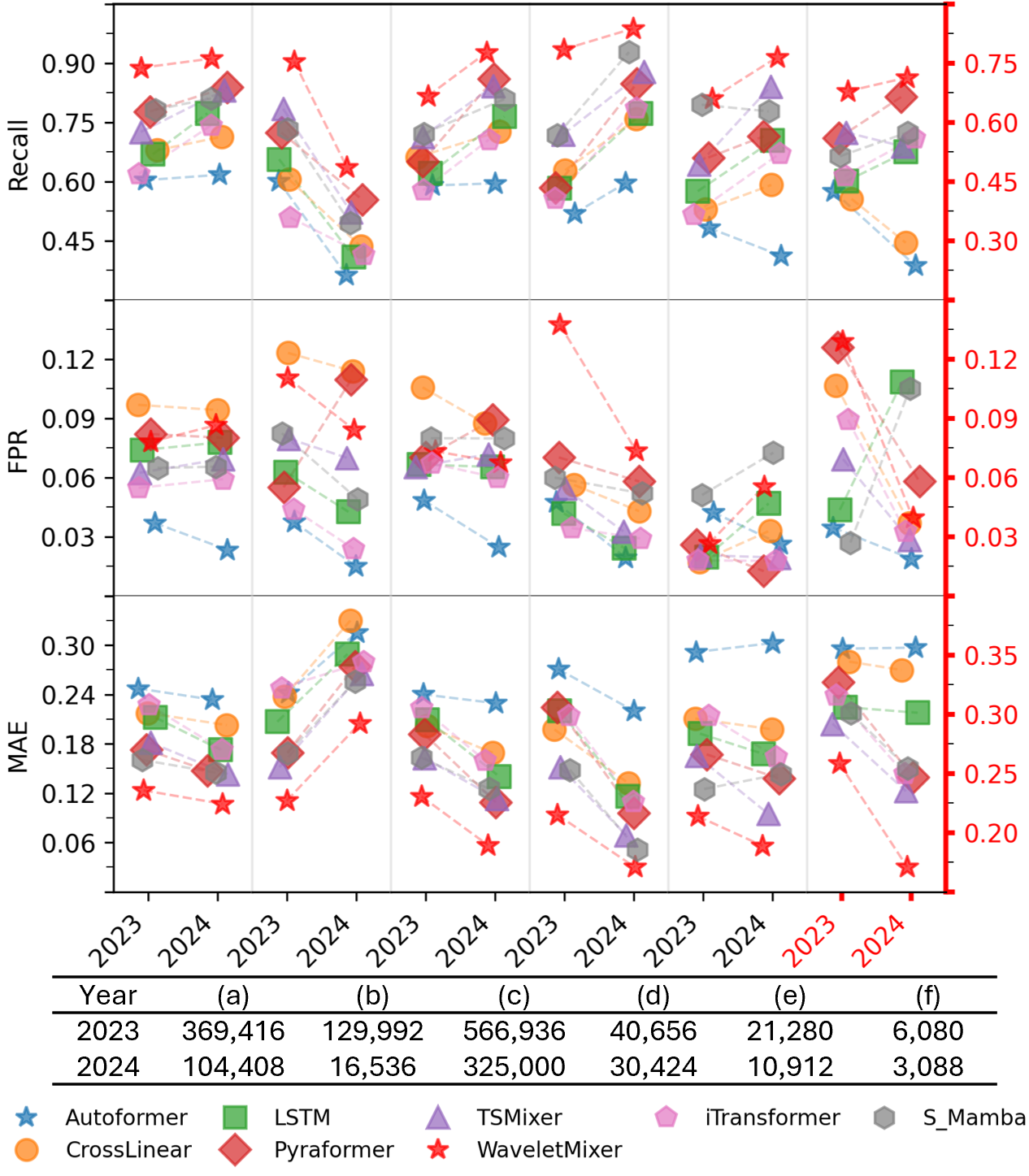}
    \caption{Model performance for the 5 dominant land-cover types in 2023 and 2024: (a) Evergreen Needleleaf Forests, (b) Mixed Forests, (c) Woody Savannas, (d) Savannas, (e) Grasslands, and (f) Others.}
    \label{fig:lulc_performancce}
\end{figure}


To further investigate the sources of the observed performance differences between the proposed model and the comparison models, we stratify the test set by both land-cover type and year. This decoupling is motivated by the fact that fuel properties, wildfire frequency, and ignition mechanisms vary substantially across land-cover types and between years \citep{gralewicz2012factors, collins2025extremely}. Accordingly, all test samples are grouped into five dominant land-cover categories, namely Evergreen Needleleaf Forests, Mixed Forests, Woody Savannas, Savannas, and Grasslands, together with an additional \textit{Others} category defined by burned area, and evaluated separately. Representative results are shown in \autoref{fig:lulc_performancce}. 

Overall, performance differences among land-cover types are substantial. For WaveletMixer in 2024, recall reaches approximately 0.99 in Savannas, while remaining near 0.7 in Mixed Forests. Correspondingly, the MSE increases from approximately 0.05 in Savannas to about 0.14 in Mixed Forests. The high recall observed in Savannas is associated with an FPR of 0.12, whereas Grasslands exhibit a much lower FPR of 0.02. These contrasts reflect underlying ecological heterogeneity. Savannas and Grasslands are characterized by relatively simple fuel structures and more linear relationships between meteorological drivers and fuel flammability, allowing the model to more readily capture changes in combustibility \citep{10.1071/WF13005, alvarado2020thresholds}. In contrast, forested systems such as Mixed Forests exhibit multilayer canopies, complex microclimates, and pronounced moisture-lag effects, which introduce nonlinear interactions between environmental drivers and fuel availability \citep{erni2020developing, collins2025extremely}. These complexities reduce predictive stability and result in lower recall and higher MSE.

\begin{figure*}[!b]
    \centering
    \includegraphics[width=\linewidth]
    {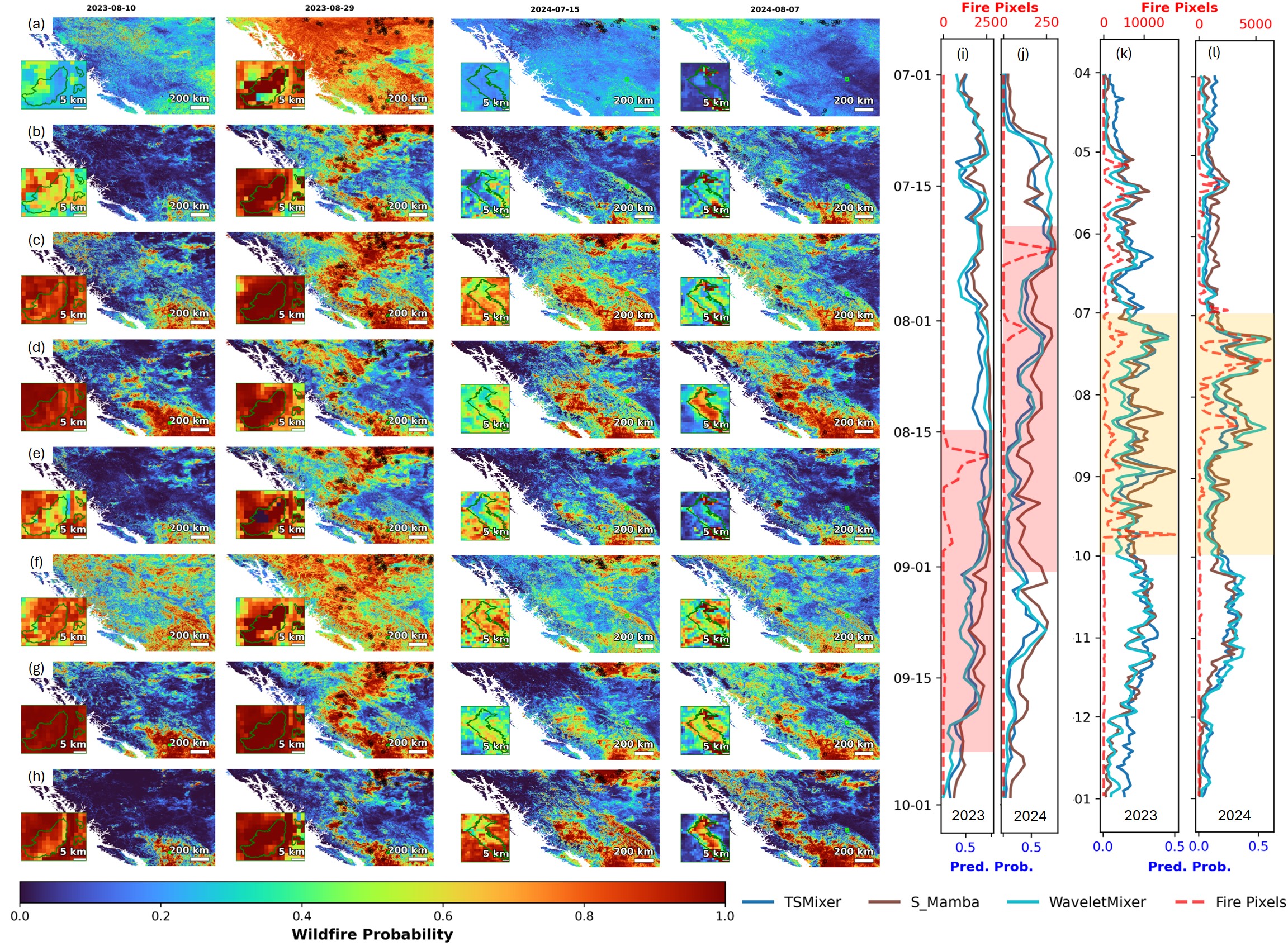}
    \caption{Spatiotemporal Evaluation of Wildfire Risk Predictions for the McDougall Creek (2023) and Jasper (2024) Fire Events and the Full Fire Season. (a)–(h) present spatial wildfire probability maps for 4 representative dates, with rows corresponding to Autoformer, CrossLinear, LSTM, Pyraformer, TSMixer, iTransformer, S\_Mamba, and WaveletMixer, respectively. (i) and (j) show event-scale temporal evolutions for the McDougall Creek Fire (2023) and the Jasper National Park Wildfire (2024). (k) and (l) show seasonal-scale temporal evolutions over the full fire season for 2023 and 2024 across the entire study area. Black open circles indicate observed fire detections. The yellow and red highlighted time intervals correspond to the time ranges of subplots (i) and (j), respectively, and to the duration of the wildfire event from its report to its containment.}
    \label{fig:qualitative_evaluation}
\end{figure*}

Within this broader context, Mixed Forests in 2024 emerge as the most challenging vegetation type for all evaluated models. Recall remains substantially lower in this class than in other land-cover types, with only limited reductions in FPR and comparatively high MSE. Specifically, in 2024, all models achieve recall values exceeding or approaching 0.6 in Evergreen Needleleaf Forests, Woody Savannas, Savannas, and Grasslands, with WaveletMixer performing particularly well in these categories and reaching recall values near or above 0.9. In contrast, recall in Mixed Forests drops below 0.6 for all baseline models, although the proposed method attains a comparatively higher value of approximately 0.7. 

In addition to fuel-related constraints in Mixed Forests mentioned above, the lower recall observed in this class can be attributed to two further factors: their fragmented spatial structure and their proximity to areas with frequent human activity \citep{erni2020developing} (see \autoref{fig:research_area}). These characteristics increase the randomness of ignition events and typically limit burn extent, thereby increasing predictive uncertainty \citep{nhess-22-3487-2022, cui2025explainable, nhess-25-4713-2025}. 


The situation differs in 2023, when the exceptionally destructive wildfire season produced large burned areas within Mixed Forests. Under these conditions, recall increased for all models and performance differences across land-cover types became less pronounced. This pattern suggests that deep learning models are able to consistently identify high-risk areas; however, whether a wildfire ultimately ignites, how it spreads, and the outcome of suppression efforts remain subject to substantial stochasticity \citep{prapas2021deep, egusphere-2025-4007}. Consequently, not all high-risk areas ultimately burn. This inherent randomness also contributed to a moderate increase in FPR for Mixed Forests in 2023 compared with 2024.

\subsubsection{Qualitative Evaluation}


Due to the black-box nature of data-driven methods, as well as disturbances introduced by data sampling, high performance metrics do not guarantee the reliability of risk prediction results \citep{li2024projecting}.To demonstrate the superiority of the proposed model over the baseline methods, we visualize representative wildfire events from 2023 and 2024, namely the McDougall Creek Fire (2023) and the Jasper National Park Wildfire (2024), by comparing model predictions approximately one week prior to ignition and during active burning (\autoref{fig:qualitative_evaluation}).

The McDougall Creek Fire, which ignited on 15 August 2023 in the Kelowna Census Metropolitan Area of British Columbia, burned approximately 13,500 hectares, triggered 208 evacuation orders, displaced nearly 35,000 residents from West Kelowna, and resulted in estimated insured losses of  \$480 million \citep{bcws2023summary, ha2025measuring}.

The Jasper Wildfire, which occurred in July 2024, was ignited by lightning and subsequently intensified by tornado-force fire-generated winds under extremely dry conditions. The event ultimately burned approximately 39,000 hectares, destroyed 358 structures, forced the evacuation of more than 25,000 residents and visitors, and caused insured losses exceeding \$880 million \citep{jasper2024wildfire, ica-adv-5-29-2025}.

From a regional perspective, except for Autoformer (\autoref{fig:qualitative_evaluation}(a)) and iTransformer (\autoref{fig:qualitative_evaluation}(f)), the remaining baseline models exhibit broadly comparable wildfire risk patterns across the four representative dates (10 August 2023, 29 August 2023, 15 July 2024, and 7 August 2024), with elevated-risk areas generally overlapping observed ignition locations (black open circles). Specifically, for the two examined dates in 2023, higher wildfire risk is primarily concentrated along the interior southern dry valleys (Okanagan Desert). For the two examined dates in 2024, high-risk areas become more spatially dispersed across the study region, extending into the northern BC–Alberta boreal forest region and adjacent sub-boreal lowland forest landscapes. By comparison, Autoformer and iTransformer show limited discriminative ability between high- and low-risk regions, often producing spatially uniform risk fields. Under these differing background conditions, the proposed WaveletMixer consistently yields more spatially concentrated predictions with clearer boundaries, fewer false positives, and improved early-warning capability across all four dates.

Specifically, prior to ignition of the McDougall Creek Fire (first column), most baseline models either exhibit delayed responses to the impending fire event or produce spatially fragmented high-risk patterns with extensive false positives, in some cases extending into the eastern mountainous regions.By contrast, the proposed WaveletMixer accurately delineates a concentrated high-risk zone within the southern dry valleys, including the Bunchgrass and Interior Douglas-fir zones, prior to ignition, thereby demonstrating genuine early-warning capability. After ignition (second column), WaveletMixer further constrains high-risk predictions to fine-fuel-dominated and climatically drier regions while effectively suppressing spurious high-risk signals in the Eastern Montane Spruce Zone, where cooler and more humid conditions limit ignition likelihood \citep{MeidingerPojar1991}. 


A similar pattern is observed for the Jasper National Park Wildfire (columns 3 and 4). Prior to ignition (15 July 2024), the proposed WaveletMixer identifies a localized high-risk area in the southern part of Jasper National Park, near Chutes Athabasca Falls, which closely aligns with the subsequent ignition location in 22 July 2024. In contrast, the baseline models fail to provide comparable early warning signals in this region.

\begin{table*}[t]
    \centering
    \caption{Ablation studies. The
 best and second-best results are highlighted in purple and green.}
    \begin{tabular}{ccccccccc}
\hline												\hline					
Settings	&	Precision $\uparrow$	&	Recall$\uparrow$	&	F1$\uparrow$	&	PR\_AUC$\uparrow$	&	ROC\_AUC$\uparrow$	&	FPR$\downarrow$	&	MSE$\downarrow$	&	MAE$\downarrow$	\\
\hline																	
w/o lulc embedding	&	\cellcolor{green!20}\textbf{94.78}	&	78.57	&	85.92	&	96.50	&	96.21	&	\cellcolor{green!20}\textbf{4.33}	&	9.65	&	14.39	\\
w/o input mdm	&	\cellcolor{blue!20}\textbf{93.73}	&	81.35	&	87.10	&	96.52	&	96.15	&	\cellcolor{blue!20}\textbf{5.44}	&	8.76	&	14.05	\\
w/o dynamic mdm	&	91.61	&	86.99	&	89.24	&	96.95	&	96.68	&	7.97	&	7.51	&	13.02	\\
w/o laplace head	&	91.83	&	\cellcolor{green!20}\textbf{88.47}	&	\cellcolor{blue!20}\textbf{90.12}	&	\cellcolor{blue!20}\textbf{97.29}	&	\cellcolor{blue!20}\textbf{97.08}	&	7.87	&	\cellcolor{blue!20}\textbf{7.02}    &	\cellcolor{blue!20}\textbf{12.07}	\\
w/o pcmixer	&	90.76	&	84.53	&	87.54	&	95.68	&	95.58	&	8.60	&	8.59	&	17.97	\\
\hline																	
WaveletMixer	&	92.29	&	\cellcolor{blue!20}\textbf{88.21}	&	\cellcolor{green!20}\textbf{90.20}	&	\cellcolor{green!20}\textbf{97.36}	&	\cellcolor{green!20}\textbf{97.15}	&	7.37	&	\cellcolor{green!20}\textbf{6.97}	&	\cellcolor{green!20}\textbf{11.81}	\\
\hline	
\hline

    \end{tabular}
    \label{tab:ablation_studies}
\end{table*}

\begin{figure*}[b]
    \centering
    \includegraphics[width=1\linewidth]{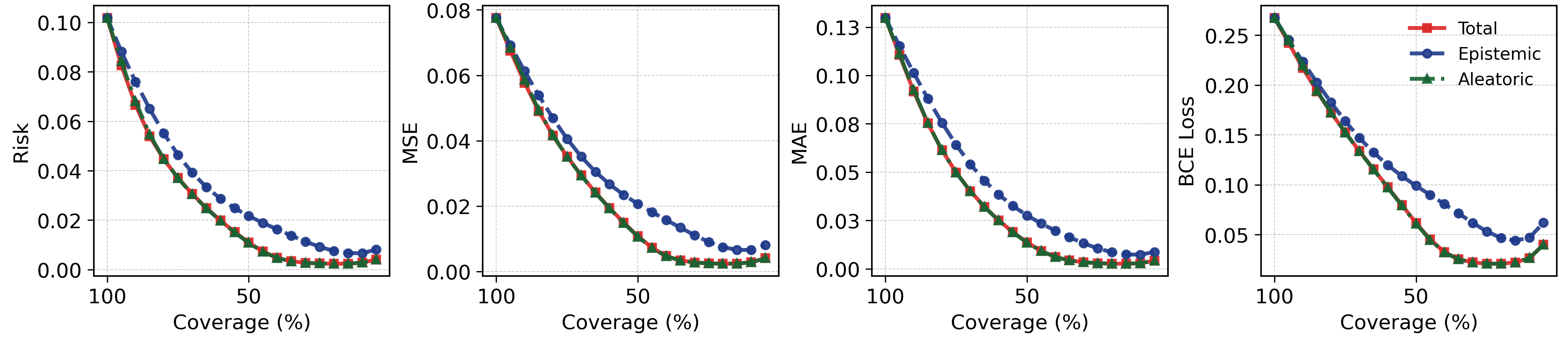}
    \caption{Uncertainty-based discard test results on the test dataset, showing Risk, MSE, MAE, and BCE Loss as functions of coverage under different uncertainty measures.}
    \label{fig:uncertainty_rejection}
\end{figure*}

After ignition (7 August 2024), most baseline models predominantly assign high risk to locations that burned on the preceding days or shortly after ignition, resulting in spatially fragmented predictions with poorly defined boundaries. By comparison, WaveletMixer consistently indicates elevated risk across the broader affected area with clearly delineated boundaries, providing a spatiotemporal risk pattern that is more consistent with the observed wildfire development.

Beyond the representative spatial evaluations for the two fire events, we further examine the temporal evolution of wildfire activity and model responses over the full burning periods of the McDougall Creek Fire and the Jasper National Park Wildfire, as well as over the entire wildfire season (from 1 April to 31 December in 2023 and 2024) across the study area (\autoref{fig:qualitative_evaluation} (i)–(l)). At the event scale (\autoref{fig:qualitative_evaluation} (i)–(j)), WaveletMixer consistently exhibits an earlier rise in predicted wildfire probability prior to the wildfire ignition, particularly for the Jasper wildfire. During the active fire phases, WaveletMixer maintains relatively stable and elevated probability levels. Moreover, during the late fire season, WaveletMixer effectively captures the subsequent decline in wildfire risk, providing a clear signal of risk attenuation as burning activity subsides.

At the seasonal scale (\autoref{fig:qualitative_evaluation} (k)–(l)), similar patterns are observed across both 2023 and 2024. While all three high-performing models broadly track the seasonal progression of fire activity, WaveletMixer produces smoother and more temporally consistent probability trajectories, with earlier increases during periods preceding major fire outbreaks and fewer spurious peaks during low-activity intervals. This behavior indicates that WaveletMixer provides a more stable and anticipatory representation of wildfire risk at both event and seasonal scales.

\subsubsection{Ablation Studies}

\autoref{tab:ablation_studies} summarizes the ablation experiments for the proposed WaveletMixer. Overall, the complete model provides the best trade-off between classification and regression metrics (F1 = 90.20, PR\_AUC $=97.36$, PRC\_AUC $=97.15$, MSE $=6.97$, MAE $=11.81$), indicating that the individual components contribute in a complementary manner to capturing multi-scale temporal dependencies, land-cover heterogeneity, and short-term transients.

Removing the LULC embedding (\textit{w/o LULC embedding} in \autoref{tab:ablation_studies}) forces the network to interpret the discrete land-cover labels as quasi-continuous values, which substantially weakens its ability to distinguish between different fuel types and fire regimes. As a result, the model becomes overly conservative, showing the highest precision (94.78~\%) and the lowest FPR (4.33~\%), but at the cost of a marked drop in recall (78.57~\%) and a clear degradation in F1, MSE, and MAE. This behaviour suggests that, without an explicit categorical embedding, the model mainly learns to trigger alarms in only a small subset of very high-risk conditions rather than generalising across ecosystems.

When the input-side multi-scale decomposition is removed (\textit{w/o input MDM} in \autoref{tab:ablation_studies}), dynamic variables are no longer pre-processed by the wavelet-based MDM block before entering the backbone. This ablation leads to a noticeable reduction in recall (81.35~\%) and F1 (87.10), together with an increase in FPR, indicating that early separation of long-term trends and short-term fluctuations helps the network to better exploit the full temporal context of the drivers. In contrast, disabling only the multi-scale branch inside the fused DDI block (\textit{w/o dynamic MDM} in \autoref{tab:ablation_studies}) produces a smaller performance drop (F1 = 89.24, MSE $=7.51$), suggesting that the input-side MDM contributes more strongly to the overall performance, while the DDI-level dynamic MDM mainly refines cross-scale interactions and moderately stabilises the error and FPR.

Omitting the Laplace transient head (\textit{w/o Laplace head} in \autoref{tab:ablation_studies}) removes the learnable exponential emphasis on the last days of the input window, effectively assigning near-uniform weights along the sequence. In this case, recall increases slightly to 88.47~\% (the highest among all  settings), but MSE and MAE are higher than in the full model and the calibration metrics (PR\_AUC, PRC\_AUC) are marginally worse. This pattern indicates that the transient head improves probability calibration and regression accuracy by focusing on recent short-term anomalies, without substantially sacrificing recall.

Finally, removing the patch-wise temporal and channel mixer in the DDI block (\textit{w/o PCMixer} in \autoref{tab:ablation_studies}) leads to the most pronounced degradation (F1 = 87.54, PR\_AUC $=95.68$, MSE $=8.59$, MAE $=17.97$). Without this component, the model lacks explicit local temporal mixing and inter-variable interaction within patches, and thus cannot fully exploit the joint response of multiple drivers. These results confirm that the PCMixer is a key element for structuring intra-patch temporal information and multi-variable dependencies, while the remaining modules provide complementary gains in multi-scale temporal modelling, land-cover specificity, and end-of-window emphasis.






\subsection{Uncertainty Evaluation}





\begin{figure*}[t]
    \centering
    \includegraphics[width=1\linewidth]{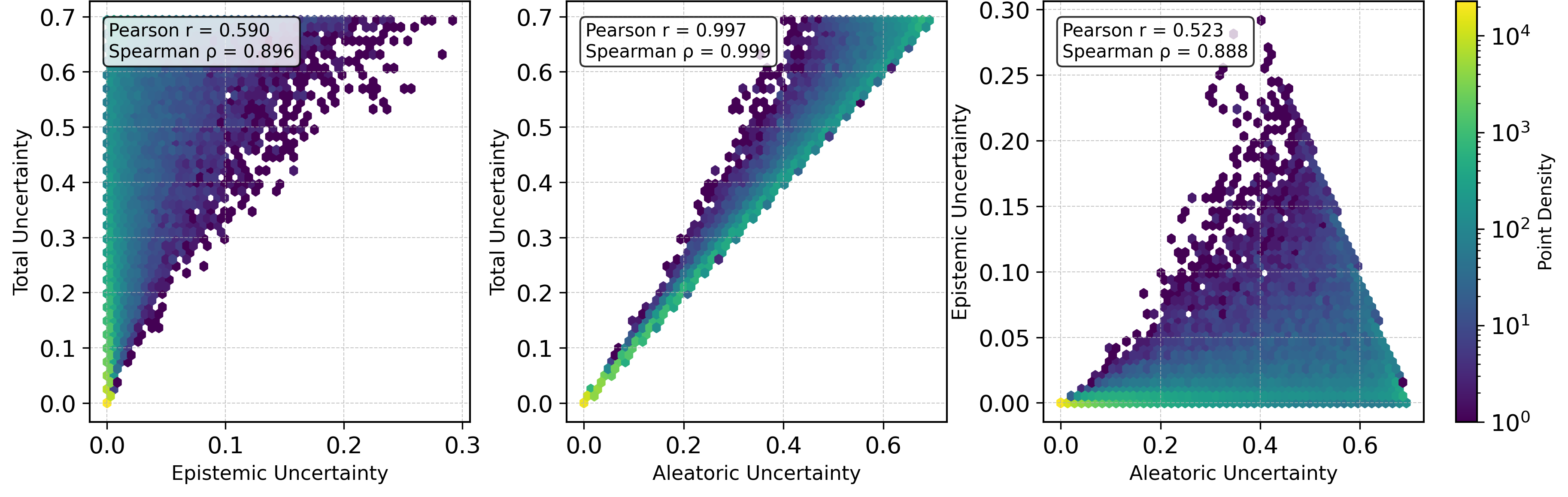}
    \caption{Correlation among different uncertainty estimation results evaluated on the test dataset.}
    \label{fig:uncertainty_correlation}
\end{figure*}


\begin{figure}[!b]
    \centering
    \includegraphics[width=1\linewidth]{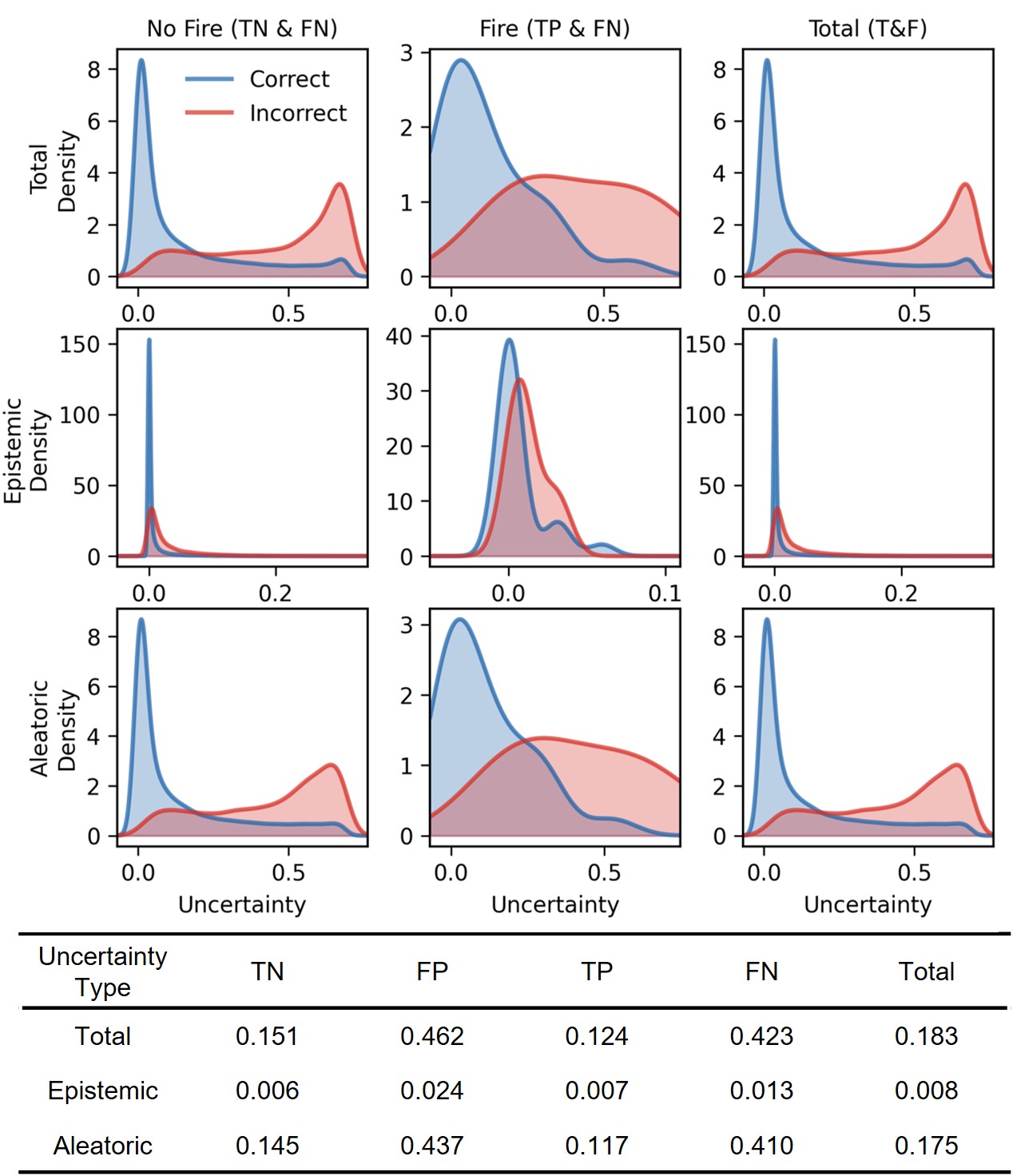}
    \caption{Distributions of total, epistemic, and aleatoric uncertainty stratified by prediction outcome and class. The kernel density estimates compare correctly and incorrectly classified samples within the No Fire (TN vs. FP), Fire (TP vs. FN), and aggregated (Correct vs. Incorrect) subsets. The table reports the mean uncertainty values for each uncertainty type across the confusion-matrix categories.}
    \label{fig:uncertainty_density_map}
\end{figure}

\begin{figure}
    \centering
    \includegraphics[width=1\linewidth]{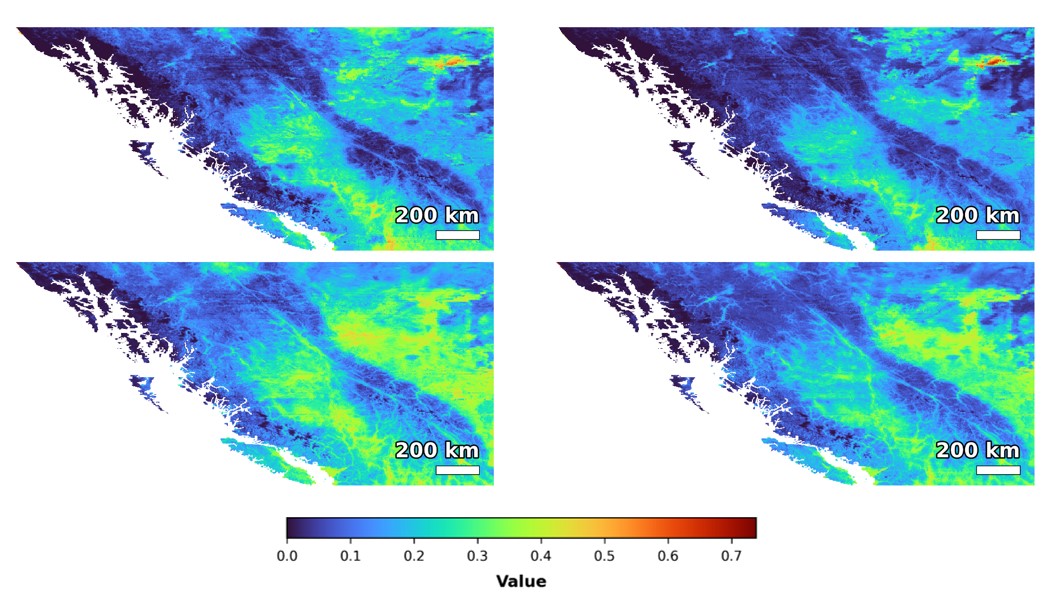}
    \caption{Spatial distributions of mean predicted wildfire probability (top row) and mean total predictive uncertainty (bottom row) for 2023 (left) and 2024 (right).}
    \label{fig:uncertainty_spatial_distribution}
\end{figure}

\begin{figure*}
    \centering
    \includegraphics[width=1\linewidth]{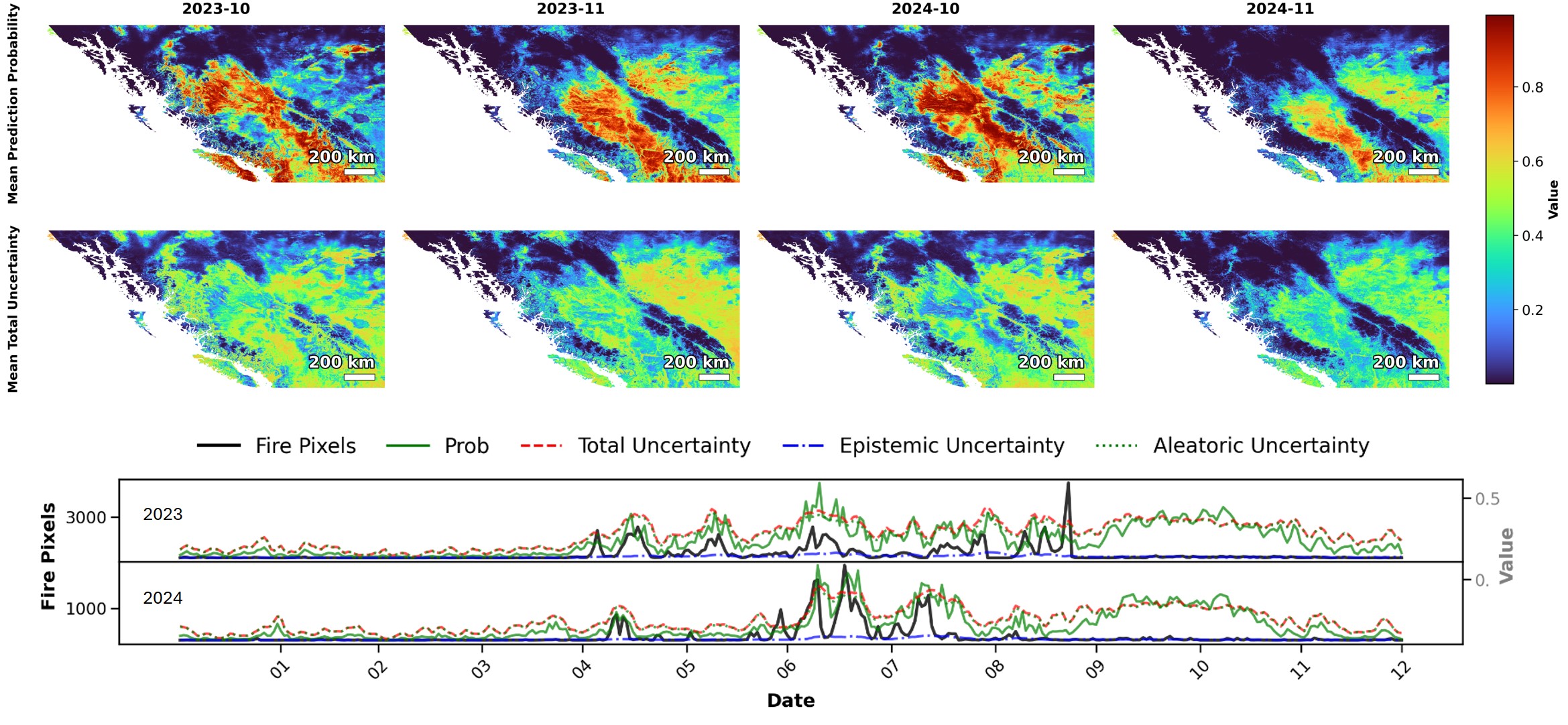}
    \caption{Spatial patterns of monthly mean wildfire risk probability and total predictive uncertainty for October and November 2023 and 2024, together with the corresponding temporal evolution of predicted probability detected fire pixels and uncertainty components.}
    \label{fig:uncertainty_temporal_distribution}
\end{figure*}

Predictive uncertainty should reflect empirical prediction errors \citep{guo2017calibration}. Following \citet{lang2022global, kondylatos2025uncertaintyawaredeeplearningwildfire, zhang2025calibration}, we evaluate uncertainty quality using uncertainty-based discard tests and uncertainty density analysis. In the discard test, predictions are progressively removed according to increasing uncertainty, and a trustworthy uncertainty estimate should yield a monotonic reduction in empirical error. As shown in \autoref{fig:uncertainty_rejection}, WaveletMixer exhibits a clear and smooth monotonic decrease across all evaluation metrics as coverage decreases, indicating that higher predicted uncertainty is consistently associated with larger prediction errors. Aleatoric uncertainty achieves a faster error reduction at low discard ratios, whereas epistemic uncertainty exhibits a more gradual improvement, reflecting their distinct roles in capturing data-level noise and model-related uncertainty, respectively.

Beyond this positive association, recent studies emphasize the importance of disentangling aleatoric and epistemic uncertainty rather than treating uncertainty as a single aggregated quantity \citep{valdenegro2022deeper, NEURIPS2024_5afa9cb1}. We therefore quantify the statistical relationship between different uncertainty components using Pearson’s and Spearman’s correlation coefficients (\autoref{fig:uncertainty_correlation}). The results indicate a moderate linear correlation but a strong rank-order association between aleatoric and epistemic uncertainty, consistent with prior findings \citep{kondylatos2025uncertaintyawaredeeplearningwildfire}. In addition, aleatoric uncertainty exhibits a substantially stronger association with total predictive uncertainty, which aligns with their highly similar discard behaviors in \autoref{fig:uncertainty_rejection} and suggests that overall predictive uncertainty is largely dominated by irreducible data variability.

Uncertainty density analysis stratified by prediction outcome further demonstrates that incorrect predictions are consistently associated with higher uncertainty than correct predictions across both fire and non-fire classes (\autoref{fig:uncertainty_density_map}). Total and aleatoric uncertainty display highly similar distributional patterns across all confusion-matrix subsets, whereas epistemic uncertainty exhibits a stronger dependence on prediction context, remaining low for non-fire samples but increasing for missed fire events. This pattern suggests that false negatives are more strongly linked to model-level uncertainty, while false positives are primarily driven by data-level variability.

Finally, spatiotemporal analysis indicates that predictive uncertainty exhibits coherent spatial and seasonal structure and does not simply mirror predicted wildfire probability. As shown in \autoref{fig:uncertainty_spatial_distribution}, regions characterized by persistently high fire susceptibility tend to exhibit relatively constrained uncertainty, whereas transitional zones display broader and more heterogeneous uncertainty patterns. In addition, late fire-season conditions are associated with elevated uncertainty even at comparable predicted probability levels (\autoref{fig:uncertainty_temporal_distribution}). Together, these patterns indicate that predictive uncertainty primarily reflects the stability of fire-risk discrimination across environmental contexts and phases of the fire season, rather than the magnitude of the predicted probability itself.

Overall, the uncertainty evaluation demonstrates that the proposed framework yields predictive uncertainty that is both informative and structured, thereby enhancing the trustworthiness of data-driven wildfire risk prediction. Uncertainty-based discard tests show a consistent monotonic reduction in empirical errors as uncertain predictions are removed, confirming that estimated uncertainty meaningfully reflects prediction reliability. Decomposition further reveals that total uncertainty is largely dominated by the aleatoric component, while epistemic uncertainty remains comparatively low and stable, indicating that residual errors are primarily driven by irreducible data variability rather than model deficiency. Stratification by prediction outcome shows that incorrect predictions are systematically associated with higher uncertainty, with distinct uncertainty signatures for different error types, supporting the diagnostic value of uncertainty for model behavior analysis. Importantly, spatiotemporal analysis reveals that uncertainty exhibits coherent spatial and seasonal structure and does not simply mirror predicted probability, but instead characterizes the stability of risk discrimination across environmental contexts and fire-season phases. 









\subsection{Compound Controls of Multiple Drivers on Fire Risk}

\begin{figure}[t]
    \centering
    \includegraphics[width=1\linewidth]{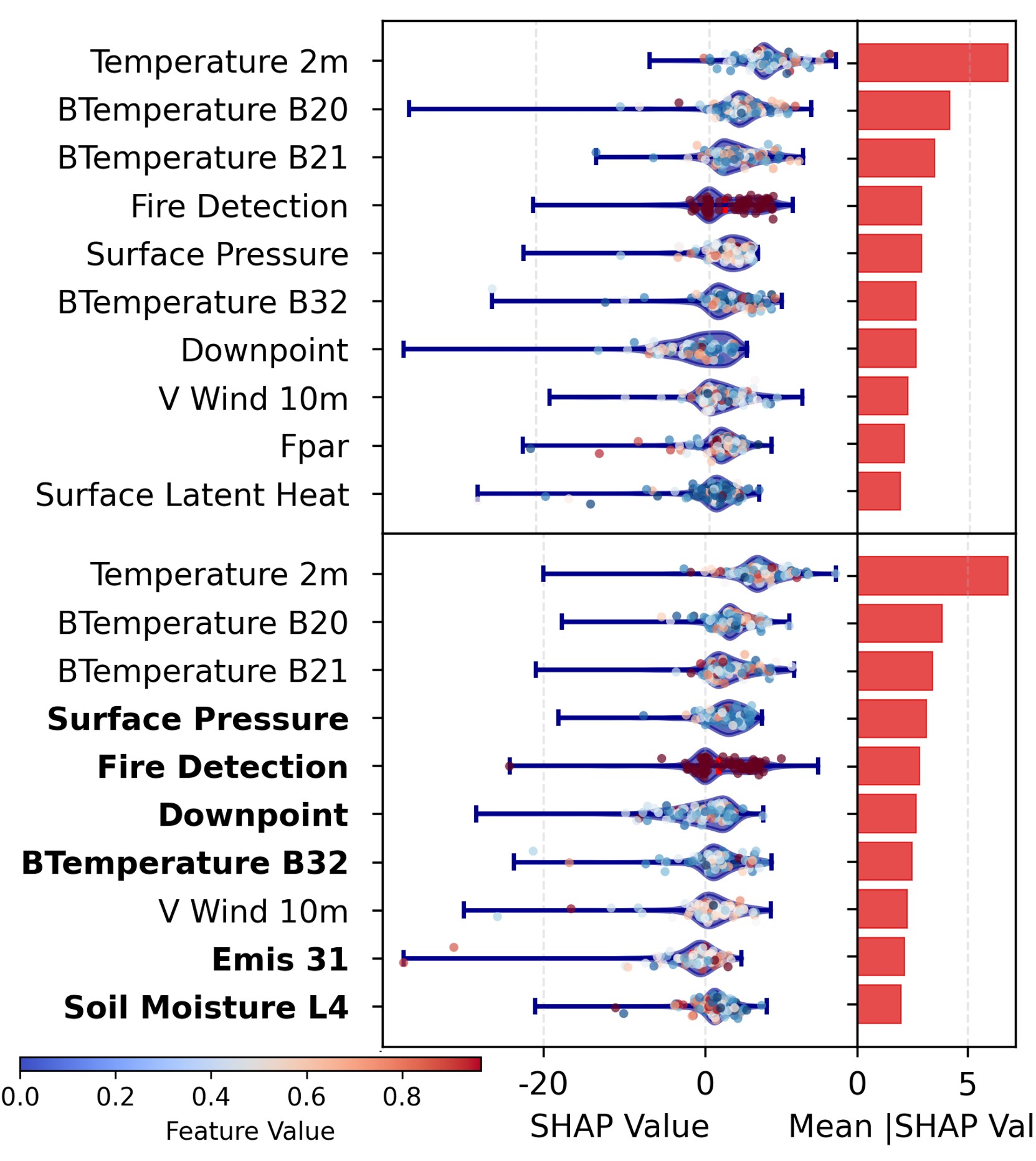}
    \caption{Global SHAP attribution for the WaveletMixer wildfire probability model in 2023 (top) and 2024 (bottom). For each year, the left panel shows the distribution of SHAP values for the top-ranked predictors, with point colors indicating the corresponding feature values, while the right panel summarizes feature importance using mean absolute SHAP values. }
    \label{fig:shap_global}
\end{figure}

\begin{figure*}
    \centering
    \includegraphics[width=1\linewidth]{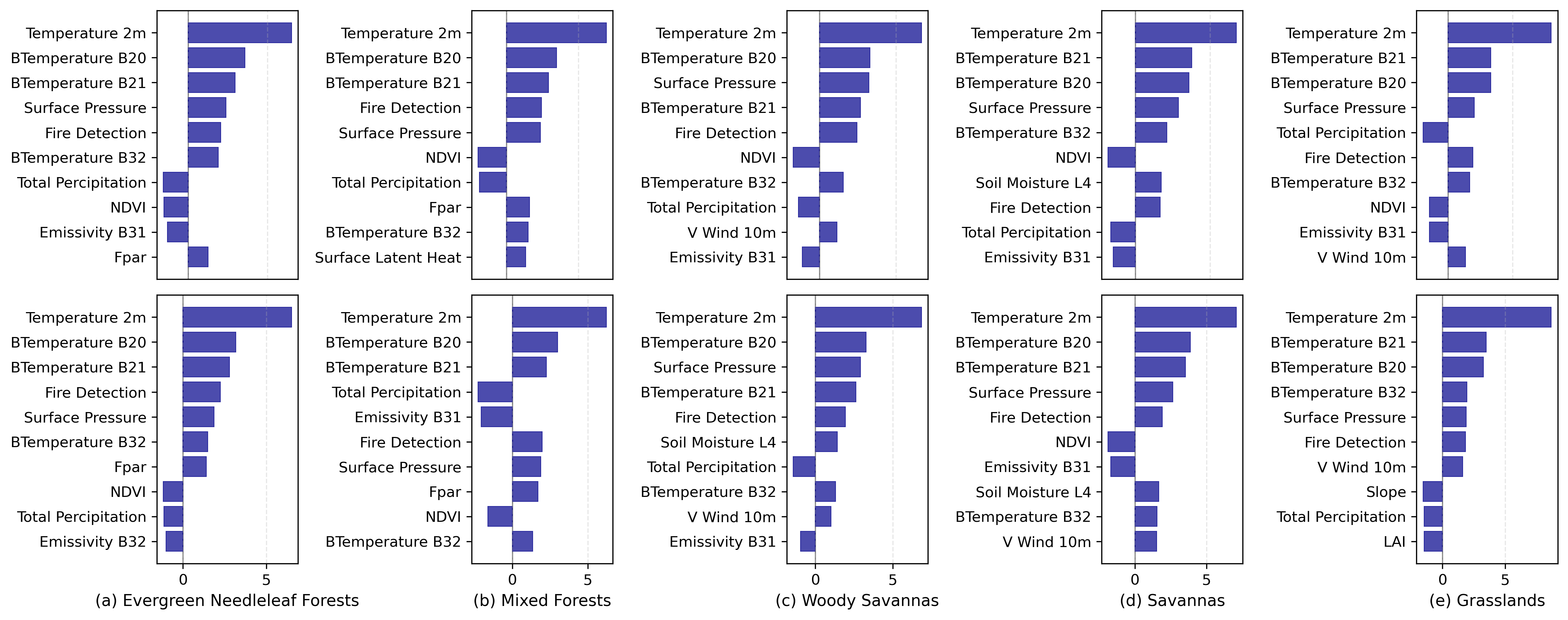}
    \caption{Land-cover-specific SHAP attribution (signed mean SHAP values) of wildfire probability for 2023 (top row) and 2024 (bottom row) across the five dominant ecosystem types in the study region: (a) Evergreen Needleleaf Forests, (b) Mixed Forests, (c) Woody Savannas, (d) Savannas, and (e) Grasslands.}
    \label{fig:shap_lulc}
\end{figure*}

Wildfire occurrence and spread across western Canada are shaped by interacting gradients in climate, topography, and fuel structure, which together regulate how meteorological forcing and surface conditions translate into ignition probability and sustained fire propagation \citep{erni2020developing,jain2024drivers}. Because these controlling factors operate differently across fire seasons and ecosystem types \citep{gallardo2016impacts, gannon2021global}, interpreting model predictions requires explicit consideration of both inter-annual variability and land-cover–specific fuel constraints.

To place the WaveletMixer predictions within a more physically interpretable framework, we conducted a SHAP-based feature attribution analysis \citep{NIPS2017_8a20a862} stratified along two complementary dimensions: (i) a regional-scale analysis for 2023 and 2024 (\autoref{fig:shap_global}), and (ii) land-cover–specific analyses for the five dominant land-cover types in the study region, namely Evergreen Needleleaf Forests, Mixed Forests, Woody Savannas, Savannas, and Grasslands (\autoref{fig:shap_lulc}) \citep{friedl2019mcd12q1}. At the regional scale, feature-wise mean absolute SHAP values quantify the overall contribution magnitude of each predictor in a given year, while signed SHAP distributions indicate whether typical feature states are associated with increased or decreased predicted fire probability. In parallel, signed mean SHAP values were aggregated within each land-cover type and year to examine how the relative importance of thermodynamic, moisture-related, and vegetation-state controls varies across fuel complexes and fire seasons. Because SHAP reflects associations learned under the joint distribution of correlated predictors, the resulting patterns are interpreted as process-consistent and hypothesis-generating rather than as definitive causal attribution.

At both the regional scale (\autoref{fig:shap_global}) and the land-cover scale (\autoref{fig:shap_lulc}), fire-probability discrimination in both years is closely linked to thermodynamic and surface-heating signals. Near-surface air temperature (2-m temperature) and thermal infrared channels (brightness temperatures B20, B21, and B32) consistently rank among the top predictors in 2023 and 2024, reflecting their direct representation of near-surface heat load and land-surface thermal state. Importantly, the influence of temperature is not limited to its direct thermal effect but also manifests indirectly through temperature-mediated atmospheric dryness and fuel desiccation processes \citep{potter2021weather, gutierrez2021wildfire, Brown2023, ivison2025unprecedented}. Together, these mechanisms regulate fuel flammability and ignition efficiency at broad spatial scales \citep{magnussen2012inter, abatzoglou2021increasing, rao2023dry, NOAA2023Climate, daniels20242023, jain2024drivers, jones2024state, Kolden2025}.

At the same time, moisture-state variables (e.g., dew point, total precipitation, Soil Moisture L4, and Band-31 emissivity), together with persistence-related signals such as fire detection, exhibit notable shifts in relative importance and sign structure between 2023 and 2024. These changes indicate that the model relies on different combinations of heat stress, atmospheric dryness, and land-surface constraints to distinguish fire-prone from less vulnerable locations across the two seasons. In 2023, widespread and persistent heat stress weakened local moisture-buffering effects, enabling extensive burning of biomass-rich fuels that are typically less flammable, particularly within forest systems dominated by Evergreen Needleleaf and Mixed Forests (\autoref{fig:lulc_performancce}) \citep{sharma2022persistent, jain2024drivers, christian2024flash}. In contrast, when such regionally coherent high-temperature conditions, associated elevated vapor pressure deficit (VPD), and severe drought were no longer present in 2024, finer-scale contrasts in moisture availability and fuel continuity played a more prominent role in determining whether fires could ignite and persist \citep{jain2021relationship, sinha2025concurrent, Kolden2025, essd-17-5377-2025}.

These fine-scale constraints manifest differently across land-cover types. As shown in \autoref{fig:shap_lulc} (a) and (b), fire-probability attribution in forests is primarily governed by temperature-related variables, with modulation by fuel state and moisture conditions such as precipitation. In contrast, wind, topography, and Soil Moisture L4 contribute only modestly. Moreover, despite substantial inter-annual differences in burned area (\autoref{fig:lulc_performancce}), the relative importance of high-contribution predictors changes little between years. Although some reordering among top-ranked features occurs, the magnitude of these changes remains limited, indicating a relatively consistent attribution structure. This pattern suggests that forest systems exhibit comparatively stable fire-risk controls, with crown-fire occurrence largely conditioned by fuel state under a shared thermodynamic background \citep{gaboriau2020temperature, banerjee2020effects}.

By comparison, as illustrated in \autoref{fig:shap_lulc} (c), (d), and (e), open and transitional land-cover types show stronger sensitivity to wind forcing, topographic effects, and deep soil moisture (Soil Moisture L4), particularly in 2024. Woody Savannas exhibit enhanced sensitivity to Soil Moisture L4, while in Savannas and Grasslands the contributions of wind speed and slope to fire-probability discrimination increase markedly relative to forested systems. These patterns are consistent with fire behavior in fine-fuel-dominated landscapes, where aerodynamic exposure, terrain, and spatial variability in fuel distribution shaped by antecedent burning and deep soil moisture conditions play a more prominent role in distinguishing fire-prone locations \citep{moritz2010spatial, tymstra2010development, hou2020observational, vinodkumar2021continental, krueger2023using, rao2023dry, jain2024drivers}.

Taken together, the SHAP analysis supports a coherent, process-oriented interpretation in which thermodynamic forcing acts as a persistent organizing influence across land-cover types, while ecosystem structure and local constraints regulate how climatic forcing is translated into realized wildfire risk. When fuels are broadly rendered combustible under anomalously high temperatures, elevated VPD, and severe drought, heat-load gradients dominate fire-probability discrimination. As wildfire risk becomes more spatially selective, however, hydrological contrasts, persistence-related signals, and spread-enabling factors play an increasingly important role. The influence of these non-thermodynamic controls varies substantially across land-cover types: forests exhibit more stable attribution patterns largely governed by fuel state, whereas in open and transitional systems, particularly when the spatial coherence of thermal forcing weakens, the interaction of fine-fuel continuity, wind forcing, and topographic effects becomes increasingly important.








\section{Conclusions}

Wildfire risk prediction is inherently challenged by the nonlinear interactions among climate, weather, fuel conditions, and human activities, as well as by the stochastic nature of ignition and fire spread. In this study, we developed a long-sequence, multi-driver wildfire risk prediction framework that explicitly captures multi-scale temporal dependencies while integrating uncertainty quantification and interpretable learning. Focusing on western Canada, the proposed approach was evaluated during the record-breaking 2023 and 2024 fire seasons, providing a stringent test under extreme and highly variable fire regimes. Although the framework is not fully Bayesian, our discard tests demonstrate that the proposed entropy-based decomposition offers a practical and effective approximation of predictive uncertainty.

Quantitative results demonstrate that the proposed model consistently outperforms existing time-series prediction methods in terms of both threshold-dependent and threshold-independent metrics, while maintaining a compact parameterization and moderate computational cost. Importantly, the proposed framework effectively identifies wildfire risk zones regardless of whether these areas ultimately experience complete ignition or sustained burning, reflecting its ability to delineate spatially coherent regions of similar fire susceptibility in a manner analogous to physics-based risk representations.

We further analyze the relationship between predictive uncertainty and model behavior through uncertainty-based discard tests, the joint distribution of uncertainty and prediction accuracy, and the spatiotemporal organization of uncertainty fields. Together, these analyses show that the explicit characterization of predictive uncertainty reveals structured spatial and seasonal patterns, indicating that uncertainty is not merely noise but reflects genuine limits in risk separability under specific environmental conditions. In particular, elevated uncertainty tends to concentrate around high-risk regions during the fire season and co-occurs with high predicted risk during the late fire season, when ignition processes become increasingly stochastic and fuel–moisture conditions evolve rapidly. These patterns highlight the effectiveness of uncertainty in delineating spatiotemporal decision boundaries and underscore the importance of uncertainty-aware risk interpretation for operational decision-making.

Interpretability analyses further provide process-level insights into wildfire controls across years and land-cover types. While temperature-related drivers consistently dominate wildfire risk in both 2023 and 2024 across all land-cover types, SHAP-based attribution reveals that moisture-related constraints play a more prominent role in shaping spatial heterogeneity and land-cover-specific contrasts in 2024, in contrast to the more uniformly hot and dry conditions of 2023. Notably, this shift is more pronounced in open ecosystems such as Savannas and Grasslands than in closed and comparatively stable forest systems. In forested ecosystems, wildfire risk is primarily governed by temperature and its downstream influence on fuel state. In contrast, open ecosystems, in addition to being influenced by temperature and fuel moisture, are more strongly modulated by non-thermodynamic factors, including topography, wind forcing, and deep-layer soil moisture.These findings emphasize that interannual variability in wildfire risk arises not only from changes in meteorological extremes but also from shifts in the relative importance of fuel moisture and broader environmental conditioning across ecosystems.

Despite these advances, several limitations remain. The current framework focuses on wildfire occurrence risk rather than explicit fire spread dynamics, and the spatial resolution is constrained by the availability of long-term, consistent driver datasets. Future work will explore tighter integration with fire behavior modeling, extension to continental and global scales, and the incorporation of additional socio-environmental constraints to further improve robustness and decision relevance. 






\bibliographystyle{elsarticle-harv} 
\bibliography{main}






\end{document}